%% file: main.tex
\documentclass{article}

\usepackage[preprint]{neurips_2026}

\usepackage{pifont}

\usepackage[utf8]{inputenc} 
\usepackage[T1]{fontenc}    
\usepackage{hyperref}       
\usepackage{url}            
\usepackage{booktabs}       
\usepackage{amsfonts}       
\usepackage{nicefrac}       
\usepackage{microtype}      
\usepackage{xcolor}         

\usepackage{amsmath}
\usepackage{amssymb}
\usepackage{mathtools}
\usepackage{amsthm}

\usepackage{multirow}
\usepackage{pgfplots}
\pgfplotsset{compat=1.18}

\usepackage[capitalize,noabbrev]{cleveref}

\usepackage{float}

\usepackage{bm}
\usepackage{colortbl}       
\usepackage[table]{xcolor} 
\usepackage{multirow}                
\definecolor{lightgray}{rgb}{.9,.9,.9}
\usepackage{svg}
\usepackage{algorithm}
\usepackage{algorithmic}
\usepackage{tikz}
\usetikzlibrary{positioning, shapes, arrows, fit, calc, arrows.meta}
\usepackage{wrapfig}
\usepackage{arydshln}
\newcommand{\vect}[1]{\bm{#1}}
\newcommand{\origin}{\overline{\bm{0}}}

\newcommand{\tb}[1]{\textbf{#1}}

\newcommand\R{\mathbb{R}}
\newcommand\eeg{^{\text{eeg}}} 
\newcommand\id{^{\text{id}}}
\newcommand\shared{^{\text{shared}}}
\newcommand\proc{_{\text{proc}}}

\title{Latte: Hyperbolic Lorentz Attention for Joint-Subject EEG Classification}

\author{%
  Ahmad Bdeir \thanks{Denotes equal contribution} \\
  Data Science Group\\
  Universität Hildesheim\\
  \texttt{bdeira@uni-hildesheim.de} \\
  \And
  Johannes Burchert \textsuperscript{*} \\
  ISMLL\\
  Universität Hildesheim \\
  \texttt{burchert@ismll.de} \\
   \And
  Tom Hanika \\
  ISMLL\\
  Universität Hildesheim \\
  \texttt{hanika@ismll.de} \\
  \And
  Lars Schmidt-Thieme \\
  ISMLL\\
  Universität Hildesheim \\
  \texttt{schmidt-thieme@ismll.de} \\
   \And
  Niels Landwehr \\
  Data Science Group\\
  Universität Hildesheim \\
  \texttt{landwehr@uni-hildesheim.de} \\
}

\begin{document}

\maketitle

\begin{abstract}


Electroencephalogram (EEG) classification plays a key role in medical diagnosis and brain–computer interfaces, but remains challenging due to low signal-to-noise ratios and high inter-subject variability. As a result, many existing approaches rely on subject-specific models, which fail to exploit shared structure in neural signals and do not generalize to unseen subjects. To address these limitations, we propose LAtte, a framework that combines Lorentz attention with a hyperbolic InceptionTime-based encoder to improve cross-subject generalization in EEG classification. The model explicitly decomposes EEG signals into a learned baseline component and task-relevant deviations, enabling more structured representation learning. To further improve robustness and adaptability, we incorporate subject-specific low-rank adaptation (LoRA) modules at both encoder and decoder levels, augmented with a Lorentz boost–based LoRA mechanism and hyperbolic projection layers to reduce overfitting in geometric representations. We evaluate LAtte with and without finetuning in three settings: subject-specific, subject-conditional, and leave-one-subject-out (LOSO) on five established EEG datasets, achieving a consistent improvement in performance over current state-of-the-art methods for smaller datasets and maintaining performance for larger datasets. Code is available at [removed during review].


\end{abstract}

\input{sections/introduction}

\input{sections/related_work}

\input{sections/background}

\input{sections/methodology}

\input{sections/experiments}

\input{sections/conclusion}

\input{sections/references}


\input{sections/appendix}



\end{document}

%% file: sections/introduction.tex
\section{Introduction}


Electroencephalogram (EEG) classification is a complex time-series task challenged by significant artifacts \citep{parbat2021novel, matt}. Specifically, ocular fluctuations (blinks/movements) and myogenic noise (muscle activity) produce large voltage swings and high-frequency interference \citep{kotte2020methods, delorme2023eeg, croft2000removal, muthukumaraswamy2013high}. These artifacts obscure neural signals and degrade the signal-to-noise ratio, hindering clinical analysis and downstream classification performance.





To address these issues, previous approaches have explored geometric methods \citep{GL}, and deep learning techniques \citep{DL_Study}. This approach was then further extended into non-Euclidean spaces due to their desirable modeling properties. Most notably, \citep{matt} introduces the Manifold Attention network (MAtt), which models the EEG data embeddings as symmetric positive-definite (SPD) matrices on a Riemannian manifold. They argue that Riemannian metrics are less sensitive to outliers and noise. More recently, \citet{bdeir2024robust} introduced HyperMAtt, which builds on the MAtt formulation by replacing the SPD operations in the decoder with Lorentzian equivalents. Our work aims to take this a step further by developing a more integrated hyperbolic approach that leverages the unique geometry in both the encoder and the decoder. 

We rely on hyperbolic spaces because they are naturally suited for representing hierarchical structures, a property that is extensively focused on in the Graph Neural Network literature \citep{mettes2023hyperbolic}. At the same time, EEG data inherently exhibits spatial hierarchies from sensor arrangements to node signal interactions. This makes EEG classification settings potentially appropriate for hyperbolic learning, which we estimate quantitatively in the paper through Gromov hyperbolicity studies. By capturing these hierarchies and modeling signal interactions directly in the Lorentz space, our approach learns more expressive latent representations and achieves good joint subject performance without requiring separately trained models for each patient. We introduce several novel components, such as the hyperbolic inception block and Lorentzian boost-based low-rank adapters (LoRAs) \citep{lora}. Specifically, the latter component is able to inject subject-specific signals through unique ID embeddings. Additionally, we combat overfitting in data-sparse problems through cut-fill augmentation and random classifier projections. This allows us to extract global EEG features while explicitly distinguishing between subject-specific distributions, resulting in improved generalization and adaptability to new subjects without retraining the model.

Our contributions in this paper are then focused on two key issues in the domain of EEG classification, namely cross-subject training and regularization for noisy input distributions:
\begin{enumerate}
    \item We propose LAtte, a hyperbolic model for EEG Classification based on hyperbolic attention and our formulation for Lorentz InceptionTime 
    \item We introduce novel subject-specific boost-based low-rank adapters.
    \item We propose a protocol for fine-tuning EEG models based on joint-subject training, and extending them to the LOSO setting. 
    \item We combat overfitting and data sparsity with hyperbolic random projection classifiers.
    \item We demonstrate the efficacy of LAtte on five well-established datasets in the domain of EEG classification. Here, we achieve an average improvement of 4.70\% in the subject-specific and 15.63\% in the subject-conditional setting over the state-of-the-art while vastly reducing total training time. Despite being a subject-dependent model, we also see a 4.73\% average improvement in the LOSO setting. 
\end{enumerate}

%% file: sections/related_work.tex
\section{Related Work}

\paragraph{EEG Classification} In the EEG classification literature, most models adopt a combination of spatial, temporal, or hybrid convolutional layers, typically followed by normalization, pooling, and a final linear classifier. Some pioneering models, such as EEGNet \citep{EEGNet} and ShallowConvNet \citep{ShallowConvNet}, both employ temporal convolutions within modular convolutional blocks. SCCNet \citep{SCCNet} extends this design by introducing spatiotemporal convolutions to facilitate spectral feature extraction. Similarly, FBCNet \citep{FBCNet} adopts an approach aligned with EEG-TCNet but integrates spectral filtering at the input stage.

EEG-TCNet \citep{EEG-TCNet} leverages causal convolutions to preserve the temporal structure of the signal, while TCNet-Fusion \citep{TCNet-Fusion} augments this framework by concatenating intermediate feature maps from the initial layers before classification. MBEEGSE \citep{MBEEGSE} represents one of the earliest applications of transformer architectures to EEG, employing EEG-specific convolutional blocks \citep{eegblock} alongside squeeze-and-excitation attention mechanisms \citep{altuwaijri2022multibranch}. MAtt \citep{matt} introduces a novel approach by replacing conventional Euclidean attention layers with manifold attention layers operating in Riemannian space.

The current research on joint-subject training is mostly limited to the task of emotion recognition and not the broader domain of EEG Classification. In \citet{li2018exploring}, the importance of different EEG features is studied, and there are some meta-learning approaches, including contrastive learning \citet{DBLP:journals/taffco/ShenLHZS23}, pre-training  \citet{DBLP:journals/sensors/CimtayE20}, and transfer learning \citet{DBLP:journals/tcyb/LiQSLH20}. More recently, \citet{burchert2024eeg} studied the joint-subject setting by introducing a training protocol with embedded subject information, addressing joint-subject generalization, and proposing the adaptation of the convolutional baselines ResNet \citep{resnettimeseries} and InceptionTime \citep{ismail2020inceptiontime} from the Time Series Classification domain.

\paragraph{Hyperbolic Approaches} Early developments in hyperbolic deep learning frequently adopted hybrid architectures that paired Euclidean encoders with hyperbolic decoders \citep{mettes2023hyperbolic}. This design choice mitigated the computational overhead associated with hyperbolic operations and circumvented the absence of well-established hyperbolic analogues for many standard Euclidean components. However, recent work has increasingly shifted towards fully hyperbolic models, motivated by the desire to better exploit the geometric inductive biases of hyperbolic space.

\citet{chen2022fully} advanced this direction by introducing hyperbolic counterparts to several foundational components, including fully connected layers, graph convolution layers, and attention mechanisms. Their architecture employs square Lorentzian distance as a similarity measure and learns class prototypes directly in hyperbolic space, using this same metric for classification loss, building on recent works such as \citep{atigh-et-al-2022}, \citep{kim2023hier}, and \citep{mettes2023hyperbolic}.


This approach was later extended to EEG tasks, most notably in the works of \citet{nguyen2025neural} and \citet{bdeir2024robust}. \citet{nguyen2025neural} proposed a unified framework for neural networks on symmetric spaces of noncompact type, which include both hyperbolic and SPD manifolds. Their approach is built around a generalized formulation of the point-to-hyperplane distance, from which they derive closed-form expressions to construct hyperbolic fully connected layers and attention mechanisms. Applying this to EEG resulted in great performance gains.
 

In parallel, \citet{bdeir2024robust} introduced HyperMatt, an extension of the Matt framework \citep{matt} that replaces the original SPD decoder with a Lorentz-based hyperbolic decoder. Specifically, they project the SPD matrices on the Lorentz space and apply Lorentzian attention and a hyperbolic MLR as a final classification layer. Although this represents an important step toward leveraging hyperbolic geometry for EEG, the model has limited joint subject performance.



%% file: sections/background.tex
\section{Background}
\paragraph{Lorentz Manifold} Hyperbolic space is a Riemannian manifold characterized by constant negative sectional curvature $-1/K < 0$ where $K$ is the curvature surrogate. In the following work, we adopt the Lorentz model of hyperbolic space (also known as the hyperboloid model), which uses the upper sheet of a two-sheeted hyperboloid within Minkowski space. The $n$-dimensional Lorentz space is then defined as $\mathbb{L}^n_K = (\mathcal{L}^n, \mathfrak{g}_{\vect{x}})$, where the manifold $\mathcal{L}^n$ is

\[
\mathcal{L}^n := \left\{ \vect{x} = [x_t, \vect{x}_s] \in \mathbb{R}^{n+1} \,\middle|\, \langle \vect{x}, \vect{x} \rangle_{\mathcal{L}} = -K,\ x_t > 0 \right\},
\]

The Lorentzian inner product defines the Riemannian metric $\langle \vect{x}, \vect{y} \rangle_{\mathcal{L}} := -x_t y_t + \vect{x}_s^\top \vect{y}_s.$
Following terminology from special relativity, we refer to $x_t$ as the \emph{time} component and $\vect{x}_s$ as the \emph{space} component. We can then use the definition of the inner product and the terminology to define the origin of the space as $\origin^K = [\sqrt{K}, 0, \dots, 0]^\top$.

\paragraph{Distance} The shortest path between two points on the manifold follows the curvature of space and is known as a geodesic. For any two points $\vect{x}, \vect{y} \in \mathbb{L}^n_K$, shortest distance between them is defined as
\[
d_{\mathbb{L}}(\vect{x}, \vect{y}) = \sqrt{K} \, \mathrm{acosh}\left(\frac{-\langle \vect{x}, \vect{y} \rangle_{\mathcal{L}}}{K}\right).
\]
The formulation for the squared distance, as proposed by \citet{law2019lorentzian}, can be calculated as
\[
d^2_{\mathbb{L}}(\vect{x}, \vect{y}) = \| \vect{x} - \vect{y} \|_{\mathbb{L}}^2 = -2K - 2 \langle \vect{x}, \vect{y} \rangle_{\mathcal{L}}.
\]
\paragraph{Exponential and Logarithmic Maps.} As a Riemannian manifold, the Lorentz space is locally Euclidean, allowing approximation at any point $\vect{x}$ using the tangent space $\mathcal{T}_{\vect{x}} \mathcal{L}$. We present the exponential and logarithmic maps that move points from and to the tangent space in \ref{operations}, along with additional Lorentzian operations.

%% file: sections/methodology.tex
\section{Methodology}

\subsection{Problem Setting} 

We are given a set of subject-specific EEG sequences and their corresponding classification labels. Assuming that the set of possible classes is fixed, the objective is to classify new EEG recordings from the same subject based on patterns in the signals. Let $x^{\text{eeg}} \in X^{\text{eeg}} := \mathbb{R}^{C \times T}$ denote an EEG recording with $C$ channels and $T$ time points,
$x^{\text{id}} \in X^{\text{id}} := \{1,\ldots,S\}$ the subject identifier, and
$y \in Y := \{1,\ldots,K\}$ the corresponding class label.

Given $N$ samples $\{(x\eeg_i, x\id_i, y_i)\}_{i=1}^N$ drawn from an unknown distribution $d$, the task is to learn a model that maps unseen EEG recordings $x\eeg$ from the same distribution to their ground-truth class $y$.

\subsection{Joint-Subject Training with LoRA}
Conventional EEG classification methods typically train separate models for each subject, motivated by the assumption that inter-subject variability in neural patterns and data distributions hinders effective joint training. However, this subject-specific approach suffers from heavier computational requirements and model storage, and does not leverage any possible cross-subject signals. And although some previous work has attempted this, they have seen limited performance improvements. 


Namely  \citet{burchert2024eeg}, attempts to learn joint models by embedding the subject ID and concatenating it with the channel dimension of the input data. However, this has several disadvantages, especially in convolutional networks, due to the limited positional interactions of the ID and the remaining EEG channel information. 
To remedy this, we directly incorporate subject ID embeddings in the data transformations by adding subject-specific low-rank adapters \citep{lora}.

\subsubsection{Lorentz Boost LoRA Layer}
\label{meth:lblora}

The LoRAs are integrated in two main components of the LAtte model, the initial Euclidean input block and the hyperbolic projection layers. Direct additive Euclidean updates do not respect the manifold geometry, and tangent-based approaches require expensive projection operations that may distort the adapted features and lead to instability. Thus, we introduce the Lorentz Boost LoRA (LB-LoRA), which re-parameterizes hyperbolic transformations along the radial direction of the manifold, enabling compact modeling of individual differences without significant parameter overhead.

Let $\vect{x}\in \mathbb{L}^n_K$, be a point on the Lorentz model. A pure Lorentz boost in the unit direction $\vect{v}$ with rapidity  $\xi$ transforms $\vect{x} = (x_t, \vect{x_s})$ as:
\begin{equation}
\text{Boost}(x; v, \xi) =
\begin{bmatrix}
x_t \cosh \xi + (\vect{x_s} \cdot \vect{v})\, \sinh \xi \\
\vect{x_s}
+ \vect{v} \left[ (\vect{x_s} \cdot \vect{v})(\cosh \xi - 1) + x_t \sinh \xi \right]
\end{bmatrix}
\end{equation}
For a subject $s$ with a LoRA rank $r$, we define the learnable parameters as directions $\{v_i^{(s)}\}_{i=1}^r$ and scalar magnitudes $\{\mu_i^{(s)}\}_{i=1}^r$. The final adapted output $y^{(s)}$ is obtained by sequentially applying these boosts to the shared output:
\begin{equation}
    y^{(s)} = \xi^{(s)} \circ \xi{r-1}^{(s)} \circ \dots \circ \xi_1^{(s)} (h_{shared})
\end{equation}
where $\xi_i^{(s)}(h) = \text{Boost}(h; \frac{v_i^{(s)}}{\|v_i^{(s)}\|}, \alpha \cdot \mu_i^{(s)})$ and $\alpha$ is a scaling hyperparameter.
This formulation ensures that all adaptations remain on the manifold without requiring artificial projection steps, providing a geometrically consistent method for subject-specific tuning.

\subsection{LAtte Model Overview}
LAtte is a hyperbolic architecture for cross-subject EEG classification that aims to separate and model subject-invariant EEG patterns and subject-specific variability. The model consists of three main stages: a subject-aware EEG processor, a hyperbolic temporal encoder, and a prototype hyperbolic decoder, as shown in \Cref{code} and \Cref{fig:latte}. 

First, raw EEG signals are processed by a Euclidean spatial and spatio-temporal input block. The choice of input block is flexible, but we rely on the implementation by \citet{altaheri2025temporal} and \cite{SCCNet}. We additionally adapt the block to use subject-specific LoRAs to explicitly inform shifts in distributions between subjects.

Second, LAtte projects the embeddings onto the hyperbolic space and applies a dual-branch hyperbolic encoder inspired by EEG baseline correction. A baseline branch with average pooling learns a smoothed reference representation, while a task branch with max pooling models the task signals. Their difference is assumed to highlight task-relevant deviations. The embeddings are then fed into a Lorentz attention module to capture long-range temporal dependencies. 

Lastly, a hyperbolic projection layer with boost-based LoRAs reduces dimensionality, and classification is performed using a Lorentzian prototype decoder based on the squared geodesic distances. The full model is trained jointly across all subjects.

\subsubsection{Subject-Conditional Preprocessor}

Our implementation follows the TCFormer multi-kernel CNN (MK-CNN) preprocessing
block~\cite{altaheri2025temporal}, which applies parallel temporal convolutions with
kernel lengths $K_c = \{20, 32, 64\}$ to capture $\beta$-, $\mu$-, and
$\theta$/$\alpha$-band activity simultaneously.
The concatenated multi-scale feature maps are then passed through a depthwise spatial
convolution over EEG channels, average pooling, a second grouped temporal convolution,
and optional grouped squeeze-and-excitation attention across kernel groups, yielding a
compact sequence of feature tokens fed into the Transformer encoder.

To address cross-subject variability without retraining the full model per subject,
we augment the two temporal convolution stages of the MK-CNN block with per-subject
LoRA adapters (Eq.~\ref{Eq:lora}).
Each temporal convolution weight $W_p$ is decomposed into a shared component
$W_p^{\shared}$, learned jointly across all subjects, and a subject-specific low-rank
correction $\Delta W_p^s = Q_p^s (R_p^s)^\top$, where
$Q_p^s \in \mathbb{R}^{I_p \times K_p}$ and $R_p^s \in \mathbb{R}^{J_p \times K_p}$
with rank $K_p \ll I_p, J_p$.
The shared weights capture the general temporal and spectral structure common to
all subjects, while the low-rank adapters model subject-specific deviations in how
EEG rhythms are expressed. One example of this is differences in the dominant frequency or
amplitude envelope of $\mu$/$\beta$ oscillations during motor imagery.
At inference time, selecting a subject's adapter only depends on their index, adding
$(I_p + J_p) K_p$ parameters per subject per adapted layer with minimal increase in
computational cost over the base forward pass.


Processed inputs $\vect{z}_n$ are mapped to the Lorentz manifold via time element concatenation and partitioned into minimally overlapping patches to capture local patterns. These patches are encoded and aggregated using Lorentzian centroid, which computes their hyperbolic average. This strategy enhances representation learning and improves efficiency by reducing effective sequence length.




\subsubsection{Baseline-Deviation Encoder}

The LAtte encoder is inspired by baseline deduction methods commonly used in EEG analysis. Traditionally, these methods leverage paired recordings where a baseline graph is generated from resting state data and a task graph is generated at task time  \citep{MAESS2016164}. Subtracting the baseline from the task graph isolates task-specific neural activity by mitigating individual variability and signal noise. However, many real-world datasets lack explicit resting state recordings, limiting the applicability of this approach.

To address this challenge, we propose a dual-branch architecture that learns an implicit baseline directly from the task data. The first branch, dubbed the baseline block, employs a Hyperbolic Inception Block with average pooling to estimate a smoothed baseline representation. In parallel, the second branch uses a Hyperbolic Inception Block with max pooling to highlight strong activation signals, which act as the task graph. We compute the deviation between these two representations by parallel transporting each task embedding along the geodesic from its corresponding baseline embedding to the origin. The goal is to produce a baseline-corrected representation that highlights task-relevant information and suppresses signal noise.



\paragraph{Hyperbolic InceptionTime Block}

To model latent representations across multiple temporal resolutions, we translate the InceptionTime architecture \cite{ismail2020inceptiontime} to the Lorentz manifold, forming the LorentzInception block. Let $\mathcal{L}^{d}$ denote the Lorentz model of hyperbolic space and let
$x \in \mathcal{L}^{d \times T}$ be an input feature map of length $T$.
Given a bottleneck width $b$, $n$ filters per branch, and kernel sizes
$\{k_1, k_2, k_3\}$, the block computes four parallel branches on the
manifold:
\begin{equation}
\begin{aligned}
\tilde{x} &= \operatorname{HConv}_{1}^{b}(x), \quad
\hat{x} = \operatorname{HPool}_{3}(x), \\
z_i &= \operatorname{HConv}_{k_i}^{n}(\tilde{x}), \quad i \in \{1,2,3\}, \\
z_4 &= \operatorname{HConv}_{1}^{n}(\hat{x}),
\end{aligned}
\end{equation}
where $\operatorname{HConv}_{k}^{c}$ is a hyperbolic 1D convolution with
kernel size $k$ and $c$ output channels, and $\operatorname{HPool}_{3}$
is a hyperbolic pooling operator with kernel $3$. The bottleneck
$\operatorname{HConv}_{1}^{b}$ is replaced by the identity when the input has a single channel. The branch outputs are merged by a Lorentz-aware
concatenation $\bigoplus$ that preserves the manifold constraint \citep{bdeir2024fully}, and the
result is passed through hyperbolic batch normalization, activation, and
dropout:
\begin{equation}
y = \operatorname{HDrop}\!\left(
        \sigma_{\mathcal{L}}\!\left(
            \operatorname{HBN_{Gyro}}\!\left(
                z_1 \oplus z_2 \oplus z_3 \oplus z_4
            \right)
        \right)
    \right),
\end{equation}
followed by a rescaling that maps $y$ back onto $\mathcal{L}^{4n}$ \cite{bdeir2024robust}. Operations inside $\operatorname{HConv}$ and $\operatorname{HBN_{Gyro}}$
are fully hyperbolic and based on the implementations by \citet{shi2026intrinsic}, ensuring that every intermediate representation remains on the Lorentz manifold.


\paragraph{Hyperbolic MaxPool} 
We introduce a distance-based Lorentzian maxpooling operator. Given a 1D window of width $k$, stride $t$, padding $u$, and dilation $\delta$, we first compute a scalar score for each embedding as the geodesic distance to the origin $o$:
\begin{equation}
s_{b,\ell} = \sqrt{K}\,\mathrm{arcosh}\!\left(-\tfrac{1}{K}\langle x_{b,\ell}, o \rangle_\mathcal{L}\right)
\label{eq:lorentz-dist0}
\end{equation}
 We then perform Euclidean maxpooling on the scalar score sequence $\{s_{b,\ell}\}_{\ell=1}^{L}$ to obtain pooled scores and their respective indices. These indices are used to gather hyperbolic vectors from the original input. No transformations are applied to the hyperbolic vectors. Intuitively, this performs the pooling operations as a selection based on geodesic radius. For each window, we select the point furthest from the hyperbolic origin, consistent with the observation that distance from origin corresponds to higher hierarchical specificity \cite{nickel2017poincare}. 

\subsubsection{Random Projection Decoder}
\label{meth:predecoder}
Following the encoder, we apply a PLFC layer~\citep{shi2026intrinsic} augmented with an LB-LoRA to reduce
dimensionality while accounting for subject-specific distribution shifts. With the boost parameter fixed to zero ($a_k = 0$) and each row $z_k$ of the spatial weight matrix normalised to unit norm ($\|z_k\| = 1$), the PLFC forward pass simplifies to a pure linear map $x_s \mapsto \langle z_k, x_s \rangle$ on the spatial components, followed by a deterministic lift back to the Lorentz manifold. We randomly initialise the shared weights $W^{\text{shared}}_{\text{PLFC}} \sim \mathcal{U}(-\sigma, \sigma)$, normalise each row to unit length, and freeze them throughout training. This simulates a random projection, borrowing intuition from the Johnson--Lindenstrauss lemma~\citep{blumrandom}, which states that high-dimensional points can be projected into a lower-dimensional space $k = O(\log n / \varepsilon^2)$ while approximately preserving pairwise distances.

By fixing $W^{\text{shared}}_{\text{PLFC}}$ with unit‑norm rows, we effectively perform a dimensionality reduction on the manifold that lowers the learnable parameter count and acts as a regulariser. To reintroduce adaptability, we augment this frozen layer with an LB-LoRA. The final classification is performed using a prototype classifier~\citep{chen2022fully} with frozen class embeddings.

\subsection{Subject-conditional Fine-tuning}
Despite challenges such as subject-specific noise and distributional shifts, learning shared structure across subjects remains a promising strategy for constructing more robust latent representations. Building on prior work of fine-tuning for EEG Classification \citep{SCCNet} and cross-subject training with subject features \citep{burchert2024eeg}, we combine both approaches and propose the use of fine-tuning for the subject-specific setting. First, we train the cross-subject model with the LoRA embedding of the subject information and then fine-tune on the individual subject, leveraging universal patterns shared across subjects while making the different distributions explicit to the model.

\subsection*{Subject-Slot Injection and Masking for Cross-Subject Generalization}

The model's reliance on training-time subject IDs for embedding permutations is problematic for generalization applications where the model is used on an entirely unseen subject. Under standard LOSO evaluation, the held-out subject $s^*$ has no corresponding trained LoRA weights. Routing test samples through subject-conditional components would expose the model to unseen embedding distributions that would harm generalization performance.

To resolve this, we reserve the subject ID $0$ as the generalization slot unassigned to any individual. We train the LoRAs for $s^*$ as a subject-agnostic generalist via stochastic subject masking. During training, each sample $i$ in a mini-batch independently undergoes instance ID replacement with probability $p$:

\begin{equation}
    \tilde{x}_{\mathrm{sub},i} =
    \begin{cases}
        0 & z_i = 1, \\
        x_{\mathrm{sub},i} & z_i = 0,
    \end{cases}
    \qquad z_i \overset{\mathrm{i.i.d.}}{\sim} \mathrm{Bernoulli}(p).
    \label{eq:subject_mask}
\end{equation}

A training instance retains the individual subject ID with a probability of $1 - p$. This preserves personalized gradient updates for all $S$ training subjects. With probability $p$, the subject ID is replaced with the generalization ID, which allows the corresponding LoRA to capture shared gradient signals from the full subject distribution, allowing it to learn a
subject-invariant structure that can be used for unseen samples. Empirically, we find $p=0.25$ to be a good masking probability.

At test time, when the model encounters the held-out subject $s^*$, we replace every test-sample index deterministically $\tilde{x}_{\mathrm{sub},i} = 0 \quad \forall\, i \in \mathcal{D}_{s^*}$. All held-out samples are then routed through the generalized slot-$0$ adapter. The stochastic masking during training trains the slot-$0$ adapter on a mixture of all subjects. This produces a fallback adapter that aims to reduce the covariate shift induced by the unseen subject. We acknowledge that alternative approaches to this exist and compare to the two direct baselines in \Cref{apx:loso_alt}, nearest-subject routing and identity-initialized LoRAs.

%% file: sections/experiments.tex
\section{Experiments}

\paragraph{Experimental Setup}

To motivate the Lorentz embedding space, we measured the Gromov $\delta$-hyperbolicity of a randomly initialized TCFormer feature space across five EEG datasets. Encoder features were consistently more tree-like than Gaussian and shuffle baselines. While the raw signals for BCI Challenge and BCIC-IV 2a were less hyperbolic initially, they became highly hyperbolic after the input block embedding. This confirms that the processor architecture and most data impose a negatively curved inductive bias, justifying the Lorentz hyperboloid as the downstream embedding space.

We compare our LAtte model against well-established EEG Classification baselines on five widely studied EEG datasets representing multiple classification problem settings. Our baselines including ShallowConvNet \citep{ShallowConvNet}, EEGNet \citep{EEGNet}, SCCNet \citep{SCCNet}, EEG-TCNet \citep{EEG-TCNet}, TCNet-Fusion \citep{TCNet-Fusion}, FBCNet~\citep{FBCNet}, MBEEGSE \citep{MBEEGSE}, InceptionTime \citep{burchert2024eeg}, MAtt \citep{matt}, CBraMod \citep{DBLP:conf/iclr/WangZLZJLL025}, HyperMatt \citep{bdeir2024robust}, EEGFormer \citep{wang2024eegformer}, ATCNet \citep{altaheri2022physics}, EEGConformer \citep{eegconformer}, MSCFormer \citep{zhao2025mscformer}, CTNet \citep{CTNet2022}, TS-SEFFNet \citep{seffnet}, BaseNet \citep{basenet}, and TCFormer \citep{altaheri2025temporal}. Furthermore, we include three cross-subject baselines: ResNetJoint, MAttJoint, and InceptionJoint \citep{burchert2024eeg}, which are adapted from their original variants using the same approach as Inceptiontime. In the main training loop, LAtte is trained on all subjects jointly. Hyperparameters were selected based on validation performance.

For the data preparation, we follow the common steps and use the same train/val/test splitting protocol as \cite{matt} for MAMEM II and BCI Challenge. Additionally, we propose a cutfill augmentation approach to help with regularization, see Appendix \ref{apx:cutfill} and perform an ablation in \ref{apx:ablations_comp}. We strictly follow \citet{altaheri2025temporal} for the remaining datasets to preserve comparability with their reported results. For all t-tests we consider the results significant if the two-tailed $P$ value is below 0.05 and extremely significant below 0.001.

\subsection{Subject-Specific and Subject-Conditional Performance}

\begin{table}[t]
    \centering
    \caption{Performance comparison on the EEG datasets BCIC-IV\,2a, MAMEM II, and BCI Challenge. We repeated the experiments 10 times on random seeds and report the average accuracy for BCIC-IV\,2a, and MAMEM II, and the AUC for BCI Challenge. We present t-test results as ** for extreme significance and * for significance.}
    \vspace{0.1em}
    \begin{small}
    \begin{tabular}{cl|ccc}
    \toprule
    & \textbf{Models}         & \multicolumn{1}{c}{\textbf{BCIC-IV\,2a}} & \multicolumn{1}{c}{\textbf{MAMEM II}} & \multicolumn{1}{c}{\textbf{BCI Challenge}} \\ \midrule
    \multirow{13}{*}{\rotatebox[origin=c]{90}{\textbf{Subject-Specific}}}
    & ShallowConvNet  & 61.84$\pm$6.39 & 56.93$\pm$6.97 & 71.86$\pm$2.64 \\
    & EEGNet         & 57.43$\pm$6.25 & 53.72$\pm$7.23 & 74.28$\pm$2.47 \\
    & SCCNet         & 71.95$\pm$5.05 & 62.11$\pm$7.70 & 70.93$\pm$2.31 \\
    & EEG-TCNet     & 67.09$\pm$4.66 & 55.45$\pm$7.66 & 77.05$\pm$2.46 \\
    & TCNet-Fusion & 56.52$\pm$3.07 & 45.00$\pm$6.45 & 70.46$\pm$2.94 \\
    & FBCNet      & 71.45$\pm$4.45 & 53.09$\pm$5.67 & 60.47$\pm$3.06 \\
    & MBEEGSE     & 64.58$\pm$6.07 & 56.45$\pm$7.27 & 75.46$\pm$2.34 \\
    & InceptionTime     &  62.85$\pm$3.21    & 62.71$\pm$2.95     & 73.55$\pm$5.08    \\
    & MAtt   \citep{matt}      & \underline{74.71}$\pm$5.01 & 65.50$\pm$8.20  & 76.01$\pm$2.28 \\
    & EEGFormer    & 68.63$\pm$2.69 & 45.21$\pm$1.92 & 79.46$\pm$3.66 \\
    & ATCNet        & 60.25$\pm$4.47 & 44.67$\pm$2.36 & \underline{80.01}$\pm$1.87 \\
    & CBraMod & 45.22$\pm$1.26 & 40.45$\pm$2.12 & 68.49$\pm$1.17 \\
    & HyperMAtt &  74.12$\pm$2.91 & \underline{68.10}$\pm$2.41 & 78.01$\pm$1.30 \\
    & TCFormer & 71.31$\pm$2.22 & 47.64$\pm$3.10 & 78.82$\pm$1.68 \\
    \rowcolor{lightgray}
    & LAtte + FT & \textbf{79.09} $\pm$2.91 & \textbf{73.47}$\pm$1.82 & \textbf{84.76}$\pm$1.55\\
    \hdashline
    \multirow{4}{*}{\rotatebox[origin=c]{90}{\textbf{SC}}}
    & ResNetJoint      &  55.54$\pm$2.72    &54.15$\pm$1.19      & 73.09$\pm$0.72       \\
    & MAttJoint      &  61.13$\pm$0.56     & 60.71$\pm$0.29     & 75.78$\pm$1.23       \\
    & InceptionJoint    &  \underline{61.38}$\pm$1.57     & \underline{66.00}$\pm$0.36      & \underline{76.13}$\pm$0.95      \\
    \rowcolor{lightgray}
    & LAtte & \textbf{78.85}$\pm$2.55 & \textbf{71.53}$\pm$1.00 & \textbf{83.8}$\pm$1.12\\
    \midrule
    & SS: Relative $\Delta$ & 5.86\%** & 8.30\%** &  5.93\%** \\
    & SC: Relative $\Delta$ & 28.46\%** & 8.38\%**  & 10.07\%**  \\
    \bottomrule
    \end{tabular}
        \label{tab:latte_MainResultsEEG}
        \vspace{-10pt}
    \end{small}
\end{table}

We report results under subject-specific (SS) and subject-conditional (SC) settings. In the SS setting, a separate model is trained for each subject, and performance is reported as the mean across subjects with pooled standard deviation over runs. For LAtte, we fine-tune the jointly trained model on each subject using only that subject’s data and ID. In the SC setting, a single model is trained jointly across all subjects using subject metadata (e.g., subject ID).

In \Cref{tab:latte_MainResultsEEG}, we compare LAtte against state-of-the-art baselines, with reported results aggregated from \citet{matt,burchert2024eeg,bdeir2024robust}. We additionally evaluate TCFormer as a recent baseline and CBraMod as a representative EEG foundation model. Hyperparameters are selected independently for each dataset using validation performance. For BCICIV-2b and HGD, we compare against the results reported in \citet{altaheri2025temporal} (\Cref{tab:tc_results}). We exclude BCICIV-2a from this comparison since the protocol in \citet{altaheri2025temporal} does not include a dedicated validation set.

In the SC setting, LAtte consistently outperforms all baselines across datasets. Unlike prior approaches relying on static subject embeddings, LAtte uses LoRA-based subject conditioning to better model subject-specific variability. LAtte also surpasses subject-specific baselines, indicating that the hyperbolic representation benefits joint-subject learning. The performance gap is smaller on HGD, likely because its larger scale reduces subject-data scarcity and weakens the benefit of shared cross-subject structure.

Foundation models such as LaBraM \citep{DBLP:conf/iclr/JiangZL24} and CBraMod \citep{cbrabod} underperform on these smaller EEG datasets (\Cref{sec:fm_forEEG}). TCFormer remains a strong baseline on BCIC-IV\,2a and BCI Challenge, but struggles on MAMEM II, where several subjects achieve near-random performance (\Cref{tab:performanceSSVEP}). In contrast, LAtte generalizes more consistently across subjects, benefiting from hyperbolic embeddings and shared subject-conditioned representations.

\begin{table}[t]
    \centering
    \caption{Performance comparison on the EEG datasets BCICIV-2b, and HGD for the subject-specific setting. FT represents finetuning. Experiments are repeated 3 times for HGD and 5 times for BCICIV-2b to match the reporting from \citet{altaheri2025temporal}. We present t-test results as ** for extreme significance and * for significance.}
    \vspace{0.1em}
    \begin{small}
    \begin{tabular}{cl|cc}
    \toprule
    & \textbf{Models}   & \multicolumn{1}{c}{\textbf{BCICIV-2b}} & \multicolumn{1}{c}{\textbf{HGD}} \\ \midrule
    & EEGNet           & 83.65$\pm$0.46     & 85.59$\pm$0.45     \\
    & ShallowConvNet   & 81.45$\pm$0.50     & 89.75$\pm$0.54     \\
    & BaseNet          & 86.11$\pm$0.40     & 93.64$\pm$0.86     \\
    & EEG-TCNet        & 86.74$\pm$0.12     & 91.83$\pm$1.8      \\
    & TS-SEFFNet       & 84.18$\pm$0.42     & 92.45$\pm$0.85     \\
    & CTNet            & 86.91$\pm$0.29     & 94.21$\pm$0.52     \\
    & MSCFormer        & 87.60$\pm$0.39     & 94.31$\pm$0.13     \\
    & EEGConformer     & 81.89$\pm$0.45     & 94.67$\pm$0.25     \\
    & ATCNet           & 86.26$\pm$0.06     & 93.65$\pm$0.33     \\
    & TCFormer         & 87.71$\pm$0.24     & \underline{96.27$\pm$0.84}     \\
    & TCFormer N=5     & -                  & 95.79$\pm$0.51    \\
    \rowcolor{lightgray}
    & LAtte            & \underline{88.87$\pm$0.17}     & 96.18$\pm$0.67     \\
    \rowcolor{lightgray}
    & LAtte  + FT      & \tb{90.01$\pm$0.32}     & \tb{97.11$\pm$0.52}     \\
    \midrule
    & Relative $\Delta$ &  2.62\%**  &  0.8\% \\
    \bottomrule
    \end{tabular}
        \label{tab:tc_results}
        \vspace{-10pt}
    \end{small}
\end{table}

\subsection{Leave One Subject Out}

To evaluate generalization to unseen subjects, we adopt the leave-one-subject-out (LOSO) protocol \citep{kunjan2021necessity}, where the model is trained on all but one subject and evaluated on the held-out subject. This procedure is repeated for each subject, and results are averaged across folds.

We compare against the LOSO results reported in \citet{altaheri2025temporal} for BCIC-IV\,2a, BCIC-IV\,2b, and HGD (\Cref{tab:tc_results}). For MAMEM II and BCI Challenge, we evaluate additional baselines and report results in \Cref{tab:LOSOResults}. We include TCFormer \citep{altaheri2025temporal} and CBraMod \citep{DBLP:conf/iclr/WangZLZJLL025}. Since CBraMod originally uses fixed subject splits, we re-evaluate it under LOSO for consistency.

All models show a substantial performance drop under LOSO due to strong distribution shifts. This degradation is particularly severe for CBraMod and TCFormer on MAMEM II. In contrast, LAtte achieves consistently stronger performance across datasets. On the BCI Challenge, both LAtte and TCFormer generalize comparatively well, though LAtte maintains a consistent advantage. For HGD, we compare to the larger variant of TCFormer used specifically for this dataset; our results are within the margin of error, which reiterates the findings from the subject-specific setting. 

\begin{table}[t]
    \centering
    \caption{LOSO performance comparison on BCICIV-2a, BCICIV-2b, and HGD. Experiments are repeated 3 times for HGD and 5 times for the others to match the reporting from \citet{altaheri2025temporal}. We present t-test results as ** for extreme significance and * for significance.} 
    \vspace{0.1em}
    \begin{small}
    \begin{tabular}{cl|ccc}
    \toprule
    & \textbf{Models}    & \multicolumn{1}{c}{\textbf{BCICIV-2a}}  & \multicolumn{1}{c}{\textbf{BCICIV-2b}} & \multicolumn{1}{c}{\textbf{HGD}} \\ \midrule
    & EEGNet            & 52.03$\pm$0.88      & 77.89$\pm$0.73    & 57.95$\pm$0.51 \\
    & ShallowConvNet    & 47.31$\pm$1.06      & 75.58$\pm$1.01    & 72.47$\pm$0.94 \\
    & BaseNet           & 56.89$\pm$0.88      & 78.61$\pm$0.57    & 68.55$\pm$1.74 \\
    & EEG-TCNet         & 55.99$\pm$0.84      & 80.56$\pm$0.20    & 60.59$\pm$1.98 \\
    & TS-SEFFNet        & 56.74$\pm$0.83      & 77.82$\pm$0.78    & 69.99$\pm$0.57 \\
    & CTNet             & 60.09$\pm$0.94      & 80.29$\pm$0.42    & 64.6$\pm$0.9   \\
    & MSCFormer         & 54.27$\pm$1.52      & 79.20$\pm$0.95    & 61.19$\pm$1.96 \\
    & EEGConformer      & 45.59$\pm$0.66      & 75.25$\pm$0.34    & 69.92$\pm$0.38 \\
    & ATCNet            & 59.66$\pm$1.27      & 80.94$\pm$0.03    & 67.42$\pm$0.24 \\
    & TCFormer          & \underline{63.00$\pm$0.60}      & \underline{81.34$\pm$0.25}    & 69.69$\pm$2.61 \\
    & TCFormer N=5         &  -  &  -   & \underline{72.83$\pm$0.25} \\
    \rowcolor{lightgray}
    & LAtte             & \tb{65.32$\pm$0.71}      & \tb{84.06$\pm$0.37}    & \tb{73.01$\pm$1.66} \\
    \midrule
    & Relative $\Delta$ &  3.68\%**  &  3.34\%** &  0.25\%\\
    \bottomrule
    \end{tabular}
        \label{tab:loso_tc}
    \end{small}
\end{table}



\paragraph{Runtime Comparison} Joint training significantly reduces total computational requirements. On BCICIV-2a, individual subject training for TCFormer \citep{altaheri2025temporal} totals 207 minutes. In contrast, LAtte trains across all subjects in 28 minutes (plus 2 minutes for fine-tuning), achieving a $\sim$7$\times$ speedup. For a direct comparison, LOSO epoch times are similar (TCFormer: 10.5s; LAtte: 9.7s), confirming that gains stem from joint training rather than per-epoch efficiency. Notably, LAtte maintains this performance despite using hyperbolic operations, which lack optimized CUDA implementations. Tests were conducted on an NVIDIA RTX 4090 GPU and AMD EPYC 7543 CPU.


%% file: sections/conclusion.tex
\section{Conclusion}

In this work, we introduced LAtte, a novel hyperbolic framework for cross-subject EEG classification that unifies Lorentzian attention with an InceptionTime-based encoder. By combining subject-specific low-rank adapters with hyperbolic representations, LAtte effectively learns shared neural patterns from subject-dependent variability. This design enables a single unified model to generalize across individuals while maintaining adaptability to new, unseen subjects. The proposed cross-subject LAtte outperforms most of its subject-specific counterparts.
Our results on five well-established datasets prove that hyperbolic geometry offers a natural and strong inductive bias for modeling the hierarchical structure inherent in EEG data, and subject-adaptive mechanisms such as LoRA can enable scalable, generalizable EEG decoding. LAtte attempts to build on previous approaches to reduce classification sensitivity to noise and become more accessible in real-world clinical applications. Some limitations remain, including testing the ability to generalize across datasets, which requires architectural modification to account for varied input lengths, and learning on similarity-based subject subsets. 

%% file: sections/references.tex
\bibliography{main}
\bibliographystyle{abbrvnat}

%% file: sections/appendix.tex
\appendix

\section{Additional Hyperbolic Operations}
\label{operations}

\paragraph{Lorentz Concatenation}
Let $z_n^{(i)} = (z^{(i)}_{n,t}, z^{(i)}_{n,v}) \in \mathcal{L}_K^{d_i}$, where $z^{(i)}_{n,t}$ and $z^{(i)}_{n,v}$ denote the time and space components, respectively. We define
\begin{equation}\label{eq:incout}
\begin{aligned}
\mathrm{HCat}\big(z_n^{(1)}, \dots, z_n^{(N)}\big)
&=
\Bigg(
\sqrt{\sum_{i=1}^{N} \|z^{(i)}_{n,v}\|^2 + \tfrac{N-1}{K}},
\; 
z^{(1)}_{n,v}, \dots, z^{(N)}_{n,v}
\Bigg).
\end{aligned}
\end{equation}

\paragraph{Exp and Log Maps}The exponential map $\mathcal{T}_{\vect{x}} \to \mathbb{L}^n_K$ projects tangent vectors onto the manifold via:
\begin{equation}
\exp_{\vect{x}}^K(\vect{z}) = \cosh(\alpha)\vect{x} + \sinh(\alpha)\frac{\vect{z}}{\alpha}, \quad \alpha = \sqrt{1/K} \, \|\vect{z}\|_{\mathbb{L}}, \label{exp}
\end{equation}
where $\|\vect{z}\|_{\mathbb{L}} = \sqrt{\langle \vect{z}, \vect{z} \rangle_{\mathcal{L}}}$. The inverse operation, the logarithmic map, is given by:
\[
\log_{\vect{x}}^K(\vect{y}) = \frac{\mathrm{acosh}(\beta)}{\sqrt{\beta^2 - 1}} (\vect{y} - \beta \vect{x}), \quad \beta = -\frac{1}{K} \langle \vect{x}, \vect{y} \rangle_{\mathcal{L}}.
\]

\paragraph{Lorentzian Centroid} 
 \citep{law2019lorentzian} derive a closed-form for a Lorentzian centroid $\bm{\mu}_\mathbb{L}$ based on the square distances. For points $\{\mathbf{x}_i\}_{i=1}^m \subset \mathbb{L}^n_K$ with weights $\bm{\nu} \in \mathbb{R}^m$, it is computed as:

\begin{equation}
\bm{\mu}_\mathbb{L} = \frac{\sqrt{K} \cdot \sum\limits_{i=1}^m \nu_i \mathbf{x}_i}{ \left\| \sum\limits_{i=1}^m \nu_i \mathbf{x}_i \right\|_{\mathcal{L}}},
\label{cent}
\end{equation}

where $\|\mathbf{z}\|_{\mathcal{L}} := \sqrt{|\langle \mathbf{z}, \mathbf{z} \rangle_{\mathcal{L}}|}$ ensures normalization to the hyperboloid surface. This closed-form solution approximates the centroid while maintaining manifold constraints.

\paragraph{Lorentz Prototype Decoder}
A strength of hyperbolic machine learning approaches is the ability to produce fine embeddings for input data while minimally distorting the hierarchical relationships between them. This leads to a better use of the ambient embedding space and better instance clustering. 
To leverage this, we utilize prototypical classification heads. 
As opposed to the MLR classifier proposed in \cite{bdeir2024fully}, prototype classifiers employ class embeddings as cluster points and outputs the results as distances to each cluster midpoint. The center with minimal distance is the most similar semantically and is used as the classified class. 
The Lorentzian prototype decoder in this work is the direct translation of the decoder proposed by \cite{wang2021cross}. We first randomly initialize N points on the hyperboloid using the Wrapped Normal Lorentz distribution in \cite{nagano2019wrapped}. We then measure similarity using the square distance as it is better defined and computationally more efficient. 

Additionally, this layer can be set to non-learnable, which forces the class centroids to remain static during the training process. This leads to better generalization in some datasets and avoids overfitting. 
We use cross-entropy as the objective function. Thus,
\begin{equation}\label{eq:ce}
  \ell(X\eeg, X\id, y; W; \mathcal{L}):= \frac{1}{N} \sum_{n=1}^N \text{cross-entropy}(y_n, \hat y_n)
   \quad \text{with } \hat y:= \text{LAtte}(X\eeg, X\id; W; \mathcal{L})
\end{equation}

\section{EEG Data Operations}
We introduce two main data operations to help with signal processing and regularization namely embedding patching and cutfill input transformations respectively.

\subsection{Patching} We split the embedded EEG sequence into windows with minimal overlap. 
Let 
\[
z_n = (z_{n,1}, \dots, z_{n,T}) \in \mathbb{R}^{T \times d}
\]
denote the processed EEG sequence. For a fixed number of windows $w$, we define the window length $L$ and stride $g$ as
\[
L := \left\lceil \frac{T}{w} \right\rceil, 
\qquad 
g := \frac{T - L}{\max(w-1,1)}.
\]
The patching operator slices $z_n$ into $w$ evenly distributed temporal windows:
\begin{equation}\label{eq:patching}
\begin{aligned}
\text{Patching}(z_n)
&:= \big(z_n^{(1)}, \dots, z_n^{(w)}\big), \\
z_n^{(i)} 
&:= \big(z_{n,\tau_i+1}, \dots, z_{n,\tau_i+L}\big)
\in \mathbb{R}^{L \times d}, \\
\tau_i 
&:= \left\lfloor (i-1)g \right\rfloor, 
\qquad i = 1,\dots,w.
\end{aligned}
\end{equation}
The patches are then reshaped to a sequence in $\mathbb{R}^{w \times L \times d}$.

\subsection{Cut and Fill} 
\label{apx:cutfill}
We propose the use of cut and fill data augmentation to prevent overfitting. In this task, from the input data, we randomly select an interval of length $l$ uniformly sampled from a range $[l_{min}, l_{max}]$. We then replace the values of the interval with the mean of the input data and predict the intermediate values. The cut-fill can be seen in \Cref{alg:cutfill}. 


    \begin{minipage}{\textwidth}
      \begin{algorithm}[H]
\caption{Apply-Cut-and-Fill}
\label{alg:cutfill}
\begin{algorithmic}[1]
\REQUIRE $x \in \mathbb{R}^{B \times 1 \times C \times T}$ 
\REQUIRE $l_{\min}, l_{\max} \in (0,1]$ 
\REQUIRE $f \in \mathbb{R}^{1 \times 1 \times C \times 1}$ 
\ENSURE $x^{\mathrm{masked}} \in \mathbb{R}^{B \times 1 \times C \times T}$, $M \in \{0,1\}^{B \times 1 \times C \times T}$, spans $\{(s_b,e_b)\}_{b=1}^B$
\STATE $x^{\mathrm{masked}} \gets x$; \quad $M \gets \mathbf{0}_{B \times 1 \times C \times T}$
\FOR{$b = 1$ {\bf to} $B$}
    \STATE $\ell_b \sim \mathrm{Unif}\big(\{\lfloor l_{\min}T\rfloor,\dots,\lfloor l_{\max}T\rfloor\}\big)$
    \STATE $s_b \sim \mathrm{Unif}\big(\{0,\dots,T-\ell_b\}\big)$; \quad $e_b \gets s_b + \ell_b$
    \FOR{$c = 1$ {\bf to} $C$}
        \FOR{$t = s_b$ {\bf to} $e_b-1$}
            \STATE $x^{\mathrm{masked}}_{b,1,c,t} \gets f_{1,1,c,1}$ 
            \STATE $M_{b,1,c,t} \gets 1$
        \ENDFOR
    \ENDFOR
\ENDFOR
\STATE \RETURN $x^{\mathrm{masked}}, M, \{(s_b,e_b)\}_{b=1}^B$
\end{algorithmic}
        \end{algorithm}
    \end{minipage}

\section{LAtte Architecture}
\input{figures/model_figure}

\begin{algorithm}[tb]
\caption{LAtte: Lorentz Attention for EEG Classification}
\small
\label{code}
\begin{algorithmic}
\REQUIRE $\mathcal{D}^\text{train} = \{(x\eeg, x\id, y_1), \ldots, (x\eeg_N, x\id_N, y_N)\}$, $\mathcal{L}^n$, number of epochs $M$
    \STATE $\begin{aligned}
    \text{Initialization}: 
    &W_{\text{att}}, W_\text{{dec}}\shared, W_{\text{out}}, Q_{\text{dec}} \sim \mathcal{U}(-\sigma, \sigma) \\
    &Q\proc \sim \mathcal{N}(\mu,\sigma)      \\
    &R\proc, R_{\text{dec}} = 0\\    
    \end{aligned}$

    \FOR{epoch $m = 1$ to $M$}
        \FOR{each training example $(x\eeg_n, x\id_n, y_n) \in \mathcal{D}^\text{train}$}
            \STATE \hspace{-1.4em} $ \begin{aligned}[t]
              z_n      &\gets \text{Processor}(x\eeg_n, x_n\id; W\proc\shared; Q\proc, R\proc)   & \\
              z_n      &\gets \text{HyperbolicProjection}(z_n; \mathcal{L}^n) &\COMMENT{Eq. \ref{exp}}\\
              z_n      &\gets \text{Patching}(z_n)                                             &\COMMENT{Eq.\ref{eq:patching}}\\
              z_{base} &\gets \text{HyperBaselineBlock}(z_n; W_{\text{IncMax}})                &\COMMENT{Eq.\ref{eq:incout}} \\
              z_{inc}  &\gets \text{HyperInceptionBlock}(z_n)                                  &\COMMENT{Eq.\ref{eq:incout}} \\
              z_n      &\gets z_{base} - z_{inc} \\
              z_n      &\gets \text{LorentzAttention}(z_n; W_\text{att})                       &\COMMENT{\citep{chen2022fully}} \\
              z_n      &\gets \text{CentroidUnpatch}(z_n)  &\COMMENT{Eq.\ref{cent}} \\
              z_n      &\gets \text{LorentzFC}(z_n, x_n\id; W_\text{dec}\shared; Q_\text{dec}, R_\text{dec}) &\COMMENT{\cref{meth:predecoder}} \\
              z_n      &\gets \text{LB-LoRA}(z_n, x_n\id; W_\text{dec}\shared; Q_\text{dec}, R_\text{dec}) &\COMMENT{\cref{meth:lblora}} \\
              \hat{y}_n &\gets \text{LorentzPrototypeDecoder}(z_n; W_\text{out})               &\COMMENT{\cref{meth:predecoder}} \\
              \ell_n &\gets \text{Cross-Entropy}(\hat{y}_n, y_n) &\COMMENT{Eq.\ref{eq:ce}}
            \end{aligned} $
        \ENDFOR
        \STATE Compute total loss: $\ell \gets \frac{1}{N}\sum_{n=1}^{N} \ell_n$
        \STATE Update parameters: $W, W\shared, Q^s_p, R^p_n$ based on $\nabla_\theta \mathcal{L}$
    \ENDFOR
    \RETURN $W, W\shared, Q, R$
\end{algorithmic}

\end{algorithm}

In \Cref{fig:latte} and \Cref{code}, we show the LAtte architecture overview. It is composed of several modular components that together form a unified, fully hyperbolic learning pipeline. A subject-aware EEG processor extracts low-level spatio-temporal features while injecting subject identity through low-rank adapters. These features are mapped to the Lorentz manifold and passed to a dual-branch hyperbolic encoder, which separates smooth baseline structure from salient task-related activity using complementary pooling strategies. The resulting representations are refined by a Lorentz attention module to capture long-range temporal dependencies. Subject-specific Lorentz Boost LoRA layers then adapt embeddings directly in hyperbolic space, enabling personalized modeling without violating manifold constraints. Finally, a hyperbolic random projection layer regularizes the representation, and a Lorentzian prototype decoder performs classification based on geodesic similarity. 

\subsection{Euclidean LoRA Layer}
This allows us to model a richer embedding of the subject information while preserving cross-subject EEG patterns. Let
\begin{align*}
     \hat y: X\eeg \times X\id \rightarrow Y
\end{align*}
be an EEG classification model, parametrized by
$P$ many parameter matrices $W_p \in \R^{I_p\times J_p}$ ($p=1{:}P$). Here, we add low-rank adapter modules to the parameter matrices $W_p$, replacing the original $W_p$, now called $W_p\shared$, by a combination of shared and subject-specific parameters:
\begin{equation}\label{Eq:lora}
\begin{aligned}
W_p &:= W_p^{\shared} + Q_p^s (R_p^s)^T, \\
W_p^{\shared} &\in \mathbb{R}^{I_p\times J_p}, \quad
Q_p^s \in \mathbb{R}^{I_p\times K_p}, \quad
R_p^s \in \mathbb{R}^{J_p\times K_p}
\end{aligned}
\end{equation}
with a lower rank $K_p \ll I_p, J_p$ for all $p$.

The model is learned as before, and the total parameter count grows from
$\sum_p I_p J_p$ to $\sum_p I_p J_p + (I_p+J_p) K_p S$.

We initialize $Q_p^s$ with a random Gaussian and $R_p^s$ with zeros. Thus, $Q_p^s{R_p^s}^T$ is zero at the beginning of training. This enables a rich and lightweight integration of the subject information in the training process because the parameters of the model $p$ are directly dependent on the subject $s$.

\paragraph{BCI-IV 2a} \citep{MI}.
The Motor Imagery (MI) BCI-IV 2a dataset, originally released as BCI Competition IV-2a (2008), is a cornerstone benchmark for motor imagery classification. It contains EEG recordings from 9 subjects, acquired with 22 Ag/AgCl electrodes over central and surrounding scalp regions at a sampling rate of 250 Hz.
The experimental task involves four motor imagery conditions: right hand, left hand, feet, and tongue. Following established preprocessing protocols, signals were down-sampled to 128 Hz, band-pass filtered to 4–38 Hz, and segmented into 4-second epochs starting 0.5 s post-cue, yielding 438 time points per trial across 22 channels.

\paragraph{BCI-IV 2b} \citep{MI}.
The second BCI-IV is a binary MI dataset (left vs. right hand) that was collected from nine subjects using three EEG channels (C3, Cz, C4) at a sampling rate of 250 Hz. Each trial has a duration of 9 seconds. Motor imagery is performed from 3–7 seconds in sessions 1–2 (4 s window) and from 3–7.5 seconds in sessions 3–5 (4.5 s window).

\paragraph{MAMEM II} \citep{SSVEP}.
The Steady-State Visual Evoked Potentials SSVEP dataset (MAMEM II, 2016) targets frequency-tagged visual responses. EEG was collected from 11 subjects using the EGI 300 Geodesic EEG System. Participants fixated on one of five flickering visual stimuli (6.66, 7.50, 8.57, 10.00, 12.00 Hz) for 5-second intervals.
Preprocessing retained 1–50 Hz activity and focused on 8 occipital channels (PO7, PO3, POz, PO4, PO8, O1, Oz, O2), corresponding to the visual cortex. Each trial was partitioned into four non-overlapping 1-second segments (125 samples per channel), producing ~500 trials per subject of 8-channel SSVEP data.

\paragraph{BCI Challenge} \citep{ERN}.
The Error-Related Negativity (ERN) dataset, released as part of the 2015 BCI Challenge\footnote{\url{https://www.kaggle.com/c/inria-bci-challenge}}, captures error-related brain responses during a P300-based spelling task. EEG was recorded from 16 subjects using 56 Ag/AgCl electrodes at 600 Hz. The task poses a binary classification problem, with a natural imbalance favoring correct responses.
Preprocessing included down-sampling to 128 Hz and band-pass filtering between 1–40 Hz. Each trial was represented by 56 channels with 160 time points, offering a challenging benchmark due to its class imbalance and inter-subject variability.

\paragraph{High Gamma Dataset} \citep{https://doi.org/10.1002/hbm.23730}.
The High Gamma Dataset (HGD) is the largest dataset used in this work. It consists of EEG recordings from 14 subjects acquired with 128 channels at a sampling rate of 512 Hz. The paradigm involves motor execution. Subjects perform one of four classes: left hand, right hand, feet, or rest. Each trial has a duration of 4 seconds, consistent with the BCIC IV-2a protocol.
Following common practice, we restrict the analysis to a 44-channel subset covering the motor cortex, apply a 4 Hz high-pass filter, and downsample the data to 250 Hz.

\paragraph{Data splitting} To reflect realistic BCI usage, we adopt a subject-specific training scheme proposed by \citep{matt} in which data is split within each subject. For BCIC-IV-2a, the first session is used for training, with one out of eight trials reserved for validation, and evaluation is performed on the second session. In the case of BCI-IV 2b each subject completed five sessions. Sessions 1–2 contain 120 trials each, while sessions 3–5 contain 160 trials each. In total, the first three sessions (400 trials) are used for training, and the final two sessions (320 trials) are reserved for testing. For MAMEM-SSVEP-II and BCI-Challenge, the first four sessions are used for training, with one out of four trials held out for validation, and testing is conducted on the remaining sessions. Lastly, for HGD each subject completed two recording sessions: the first session, comprising approximately 880 trials, is used for training, while the second session, with around 160 trials, is reserved for testing.

\section{Gromov \texorpdfstring{$\delta$}{δ}-Hyperbolicity Analysis}
\label{apx:hyperbolicity}

\subsection{Background}

A metric space $(X, d)$ is $\delta$-hyperbolic if, for every four points
$x, y, z, w \in X$~\cite{Gromov1987}:
\begin{equation}
  d(x,y) + d(z,w) \;\leq\;
  \max\!\big(d(x,z) + d(y,w),\; d(x,w) + d(y,z)\big) \;+\; 2\delta .
  \label{eq:gromov_four_point}
\end{equation}
Rearranging, the \emph{local} $\delta$ for a given quadruple is
$\delta(x,y,z,w) = (S_1 - S_2)/2$, where $S_1 \geq S_2 \geq S_3$ are the three
sums of opposite-side pairwise distances.  Trees are $0$-hyperbolic and the two-dimensional Lorentz model has
$\delta = \log(1 + \sqrt{2}) \approx 0.88$ in absolute terms, but its diameter
is infinite, yielding an effective scale-invariant $\delta_{\mathrm{rel}}$ of
approximately $0.144$ when clipped at $10^{-5}$ from the
boundary~\cite{khrulkov2020hyperbolic}.  For comparison, a 2D sphere yields
$\delta_{\mathrm{rel}} \approx 0.97$.

Following~\cite{khrulkov2020hyperbolic}, we report the \textbf{scale-invariant
relative delta}:
\begin{equation}
  \delta_{\mathrm{rel}}(X)
  \;=\; \frac{2\,\hat{\delta}(X)}{\mathrm{diam}(X)}
  \;\in\; [0, 1],
  \label{eq:delta_rel}
\end{equation}
where $\hat{\delta}(X)$ is estimated as the 95th percentile of $\delta$ over a
large random sample of quadruples (the max is retained for reference but it is an unstable estimator that is sensitive to
outliers).

\textbf{Relevance to the Lorentz model.}  The Lorentz model $\mathbb{L}^n$ and
the Poincar\'{e} ball are isometric; hence the $\delta$-hyperbolicity constant,
the effective $\delta_{\mathrm{rel}}$, and the curvature estimation formula
from~\cite{khrulkov2020hyperbolic} transfer without modification.  Given an
estimated $\delta_{\mathrm{rel}} = \delta_X$ for a dataset, the curvature of the
target Lorentz hyperboloid can be set as
$c(X) = \bigl(0.144 \,/\, \delta_X\bigr)^2$, where $c(X)$ is the absolute value
of the negative curvature (i.e., $\langle \vect{x}, \vect{x} \rangle_{\mathcal{L}} = -1/c(X)$).

\subsection{Experimental Setup}

We extract features from the untrained input processor backbone to isolate the architectural inductive bias from learned representations.  Five EEG datasets spanning motor imagery and error monitoring are used (Table~\ref{tab:dataset_2}).

\begin{table}[htbp]
  \centering
  \caption{EEG datasets used in the $\delta$-hyperbolicity analysis.}
  \label{tab:dataset_2}
  \small
  \begin{tabular}{l l c c}
    \toprule
    \textbf{Dataset} & \textbf{Task} & \textbf{Classes} & \textbf{Trials used} \\
    \midrule
    MAMEM        & SSVEP                  & 5 & 3{,}300 \\
    BCICha       & Error monitoring       & 2 & 2{,}880 \\
    BCIC\,IV\,2a & Motor imagery          & 4 & 2{,}592 \\
    BCIC\,IV\,2b & Motor imagery          & 2 & 3{,}680 \\
    HGD          & Motor imagery          & 4 & 5{,}000 \\
    \bottomrule
  \end{tabular}
\end{table}

For each dataset, $N = 2{,}500$--$5{,}000$ trials are passed through the
randomly initialised encoder and flattened, yielding feature vectors of
dimension $D$ (ranging from $96$ to $1{,}488$ after the encoder, vs.\
$1{,}000$--$44{,}000$ for raw signals).  Pairwise Euclidean distances are
computed, and $\delta$ is estimated from $50{,}000$ random quadruples, repeated
across $n_{\mathrm{seeds}} = 5$ independent samplings for variance estimation.

\paragraph{Baselines}
To distinguish genuine tree-likeness from artifacts of high dimensionality or
marginal distributions, we compare against two structure-destroying null models:

\begin{itemize}
  \item \textbf{Gaussian baseline.}  Independent Gaussian samples
    $\vect{z} \sim \mathcal{N}(\vect{zero}, \operatorname{diag}(\boldsymbol{\sigma}^2))$
    where $\sigma_j = \operatorname{std}(X_{:,j})$.  Controls for per-dimension
    variance.

  \item \textbf{Shuffle baseline.}  Each feature column is independently
    permuted across trials, exactly preserving every marginal distribution
    (including ReLU-induced sparsity and heavy tails) while destroying all
    inter-point geometric structure.  This is the more stringent test.
\end{itemize}

The primary metric is the difference in normalised $\delta$:
\begin{equation}
  \Delta_{q95}
  \;=\; \delta_{\mathrm{rel}}^{q95}(\text{real})
    \;-\; \delta_{\mathrm{rel}}^{q95}(\text{baseline}) .
  \label{eq:delta_diff}
\end{equation}
Negative $\Delta_{q95}$ implies the real features are \emph{more} tree-like
than the structureless null.  We report $\|\Delta\|/\sigma$, the magnitude of
the difference in units of its across-seed standard deviation, as a
signal-to-separation ratio; $\|\Delta\|/\sigma > 2$ corresponds (under a normal
approximation) to $p < 0.05$.

The identical pipeline is also applied to the raw flattened EEG signals
(no encoder), providing a ``no-encoder'' control.

\subsection{Results}

Table~\ref{tab:overall} reports the overall $\delta$-hyperbolicity results.

\begin{table}[htbp]
  \centering
  \caption{Gromov $\delta$-hyperbolicity of untrained TCFormer encoder features.
    $\delta_{\mathrm{rel}}^{q95}$ is the diameter-normalised 95th-percentile
    $\delta$ (lower = more tree-like).  $\Delta_{q95}$ is the difference
    vs.\ baseline; negative values indicate stronger hyperbolicity than expected
    by chance.  $\|\Delta\|/\sigma$ quantifies statistical separation.}
  \label{tab:overall}
  \small
  \begin{tabular}{l c c c c c c}
    \toprule
    \textbf{Dataset} & $D_{\mathrm{enc}}$ & \textbf{Baseline}
      & $\delta_{\mathrm{rel}}^{q95}$ (Real)
      & $\delta_{\mathrm{rel}}^{q95}$ (Baseline)
      & $\Delta_{q95}$
      & $\|\Delta\|/\sigma$ \\
    \midrule
    MAMEM  & 1488 & Shuffle  & 0.0038 & 0.0116 & $-$0.0078 & 11.1 \\
           &      & Gaussian & 0.0042 & 0.0222 & $-$0.0180 & 352.5 \\
    \addlinespace
    BCICha & 96   & Shuffle  & 0.0238 & 0.0388 & $-$0.0150 & 26.5 \\
           &      & Gaussian & 0.0239 & 0.0565 & $-$0.0326 & 102.5 \\
    \addlinespace
    BCIC\,IV\,2a & 816 & Shuffle  & 0.0167 & 0.0193 & $-$0.0026 & 8.9 \\
           &      & Gaussian & 0.0165 & 0.0281 & $-$0.0116 & 79.5 \\
    \addlinespace
    BCIC\,IV\,2b & 960 & Shuffle  & 0.0085 & 0.0150 & $-$0.0065 & 18.2 \\
           &      & Gaussian & 0.0111 & 0.0268 & $-$0.0157 & 170.8 \\
    \addlinespace
    HGD    & 816  & Shuffle  & 0.0028 & 0.0122 & $-$0.0093 & 28.7 \\
           &      & Gaussian & 0.0036 & 0.0302 & $-$0.0266 & 143.7 \\
    \bottomrule
  \end{tabular}
\end{table}

\begin{figure}
    \centering
    \includegraphics[width=1\linewidth]{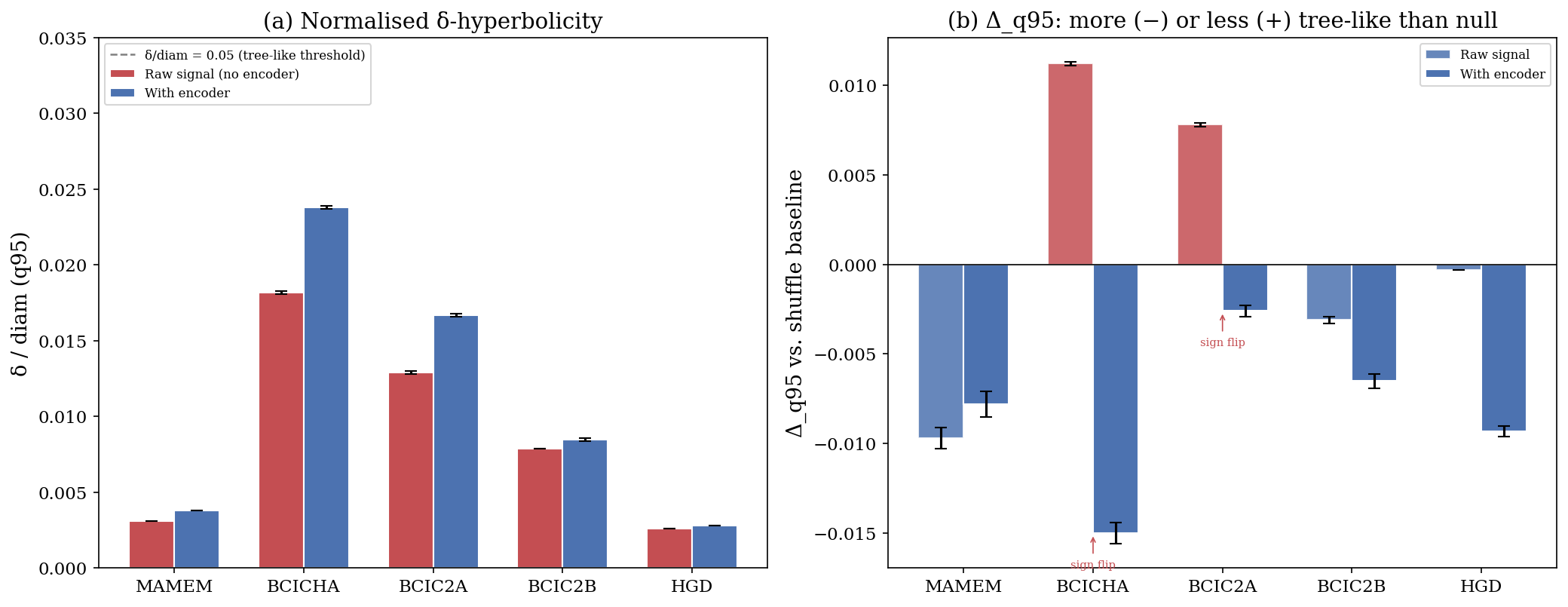}
    \caption{A graphical representation of the normalized hyperbolicity values and deviation from baseline for RAW and Encoded data.}
    \label{fig:hypergraphs}
\end{figure}
Across all datasets and both baselines, $\Delta_{q95} < 0$ with
$\|\Delta\|/\sigma$ ranging from $8.9$ to $352.5$.  The absolute
$\delta_{\mathrm{rel}}^{q95}$ values ($0.0028$--$0.0239$) are substantially
below the effective Lorentz reference of $0.144$
in~\cite{khrulkov2020hyperbolic}.  The effect is robust even under the shuffle
baseline, ruling out explanations based on sparsity or marginal shape.

\begin{table}[htbp]
  \centering
  \caption{Encoder vs.\ no-encoder comparison (shuffle baseline).
    Negative $\Delta_{q95}$ = more hyperbolic than baseline.}
  \label{tab:vs_raw}
  \small
  \begin{tabular}{l c c c c c}
    \toprule
    \textbf{Dataset} & $D_{\mathrm{raw}}$ & Raw $\Delta_{q95}$
      & $D_{\mathrm{enc}}$ & Encoder $\Delta_{q95}$
      & \textbf{Geometric change} \\
    \midrule
    MAMEM  & 1{,}000 & $-$0.0097 & 1{,}488 & $-$0.0078 & Remains hyperbolic \\
    BCICha & 8{,}960 & \textbf{$+$0.0112} & 96   & \textbf{$-$0.0150} & Euclidean $\rightarrow$ Hyperbolic \\
    BCIC\,IV\,2a & 22{,}000 & \textbf{$+$0.0078} & 816 & \textbf{$-$0.0026} & Euclidean $\rightarrow$ Hyperbolic \\
    BCIC\,IV\,2b & 3{,}375 & $-$0.0031 & 960  & $-$0.0065 & Strengthened \\
    HGD    & 44{,}000 & $-$0.0003 & 816  & $-$0.0093 & Strengthened \\
    \bottomrule
  \end{tabular}
\end{table}

As we can see in \cref{fig:hypergraphs}, and \cref{tab:vs_raw}, BCICha and BCIC\,IV\,2a produce raw EEG features that are
less hyperbolic than their shuffle baselines ($\Delta_{q95} > 0$),
consistent with the high-dimensional Euclidean concentration phenomenon. The TCFormer encoder flips
the sign of $\Delta_{q95}$ for both. For the
remaining three datasets, the encoder preserves or strengthens the existing hyperbolic
tendency.

\subsection{Hyperparameters}

The hyperparameter search for the learning rate \{$1e^{-3},1e^{-4}$\}, weight decay, \{$1e^{-1},1e^{-2},5e^{-2},1e^{-3}$\}, patching windows \{1,2,3,4\} and learning rate of the LoRA weights in the decoder \{$1e^{-1},1e^{-3},1e^{-5}$\} was repeated on random seeds and the optimal values selected on the best validation accuracy for BCI-IV 2a and SSVEP and the best validation AUC for ERN. For CBraMod and TCFormer, we extended the weight decay search grid by adding $1e^{-4}$ as well. 


\begin{table}[t]
    \centering
    \caption{Performance comparison for LOSO protocol. The experiments are repeated 5 times on random seeds, and we report accuracy for BCI-IV 2a and MAMEM II, and AUC for BCI Challenge.}
    \vspace{0.1em}
    \label{tab:LOSOResults}
    \begin{small}
    \begin{tabular}{l|cc}
    \toprule
    \textbf{Models}         & \multicolumn{1}{c}{\textbf{MAMEM II}} & \multicolumn{1}{c}{\textbf{BCI Challenge}} \\ \midrule
    EEGFormer & 29.55$\pm$5.40 & 73.30$\pm$3.55\\
    ATCNet & \underline{50.31$\pm$3.82} & 73.58$\pm$4.32\\
    CBraMod &  43.58$\pm$2.63 & 66.70$\pm$299 \\
    TCFormer & 42.18$\pm$3.91 & \underline{74.28$\pm$4.10} \\
    \rowcolor{lightgray}
    LAtte  &  \textbf{57.86}$\pm$3.66 & \textbf{75.31}$\pm$3.17 \\    
    \midrule
    Relative $\Delta$ &  15.00\%  &  1.38\% \\
    \bottomrule
    \end{tabular}
    \end{small}
\end{table}

\subsection{Per Subject Performance}
\begin{table*}[t]
\centering
\caption{Performance Comparison on the MAMEM II dataset for the subject-specific case. We report the average accuracy over 10 runs, respectively.}
\label{tab:performanceSSVEP}
\begin{tabular}{lcccc}
\toprule
\textbf{Subject} & \textbf{InceptionTime} & \textbf{MAtt} & \textbf{HyperMAtt} & \textbf{LAtte} \\
\midrule
1 & $80.40\pm2.06$ & $81.60\pm2.87$ & $\underline{89.94}\pm1.29$ & $\textbf{90.00}\pm2.18$ \\
2 & $86.60\pm1.62$ & $89.40\pm1.36$ & $\underline{90.31}\pm0.96$ & $\textbf{94.50}\pm4.62$ \\
3 & $61.60\pm3.07$ & $58.20\pm5.64$ & $\underline{68.05}\pm3.06$ & $\textbf{74.20}\pm0.16$ \\
4 & $25.00\pm4.00$ & $20.60\pm3.88$ & $\underline{30.00}\pm0.58$ & $\textbf{42.67}\pm1.89$ \\
5 & $25.00\pm6.72$ & $26.40\pm4.80$ & $\underline{30.96}\pm1.53$ & $\textbf{53.20}\pm2.34$ \\
6 & $79.20\pm1.72$ & $79.00\pm2.68$ & $\underline{80.90}\pm1.55$ & $\textbf{86.40}\pm0.70$ \\
7 & $69.20\pm1.72$ & $66.00\pm2.19$ & $\underline{69.08}\pm2.08$ & $\textbf{78.60}\pm2.07$ \\
8 & $23.60\pm1.74$ & $23.80\pm2.71$ & $\textbf{29.04}\pm0.58$ & $\underline{25.20}\pm2.03$ \\
9 & $79.40\pm2.58$ & $88.20\pm2.04$ & $\textbf{93.02}\pm3.61$ & $\underline{91.80}\pm2.84$ \\
10 & $68.60\pm3.72$ & $70.60\pm4.54$ & $\underline{75.11}\pm2.00$ & $\textbf{76.80}\pm1.38$ \\
11 & $91.20\pm2.48$ & $90.20\pm1.47$ & $\underline{92.96}\pm1.15$ & $\textbf{94.80}\pm2.57$ \\
\midrule
Summary & 62.71$\pm$2.95 & 63.90$\pm$1.95 & 68.12$\pm$1.91 & 73.47$\pm$1.82 \\
\bottomrule
\end{tabular}
\end{table*}

In \Cref{tab:performanceSSVEP}, we show the performance of our fine-tuned model LAtte and compare it to recent baselines. Here, the EEG common subject variability can be observed, where subject 3 defaults to near random performance while subject 2 achieves almost perfect accuracy. This trend is persistent for all models. However, on previously challenging subjects, e.g., 4 and 5, LAtte, with its richer subject information, is capable of leveraging features that are shared across the dataset to increase performance.

\subsection{Ablation Studies}

\subsubsection{Foundation Models for Short EEG Sequences}\label{sec:fm_forEEG}

In this ablation study, we investigate whether foundation models pretrained on large EEG corpora can improve classification on the small datasets with short sequence lengths used throughout this thesis. Specifically, we extend CBraMod by pairing its pretrained backbone with more expressive decoder architectures to better leverage the learned representations.

\paragraph{CBraMod + InceptionTime.} InceptionTime has proven effective for the BCI-IV 2a, MAMEM II, and BCI Challenge datasets in~\citet{burchert2024eeg}. Here, we use the CBraMod backbone and replace its classification head with InceptionTime layers. We evaluate two configurations: (i)~\emph{IncShallow}, using a single Inception layer, and (ii)~\emph{IncDeep}, using the standard three-layer configuration.

\paragraph{CBraMod + LAtte.} To incorporate representations learned from the large EEG corpus in~\citet{obeid2016temple}, we replace the Baseline Lorentz Inception block in our LAtte architecture with the CBraMod backbone. Since CBraMod operates on fixed patch sizes designed for long clinical recordings, its output dimensionality does not directly match the temporal resolution expected by the LAtte processor. We therefore introduce a lightweight coupling layer to bridge this mismatch:
\begin{equation}
    Z = \text{AvgPool}\!\left(\text{ReLU}\!\left(\text{Conv1D}(\mathbf{X};\, k{=}3)\right);\, T_{\text{out}}\right),
\end{equation}
where $X \in \mathbb{R}^{B \times C \times T}$ is the raw EEG input, the $\text{Conv1D}$ with kernel size $k{=}3$ and same-padding projects to the target channel dimension $d_{\text{proj}}$, and $\text{AvgPool}$ adaptively downsamples the temporal axis to $T_{\text{out}} = T_{\text{Inception}}$ to match the processor's expected sequence length. The resulting embeddings $Z \in \mathbb{R}^{B \times d_{\text{proj}} \times T_{\text{out}}}$ from the CBraMod backbone are projected onto the manifold and then replace the baseline branch, from which the output of the Lorentz Inception block is subtracted as in the standard LAtte architecture.

\begin{table}[t]
    \centering
    \caption{
Performance comparison of foundation model configurations on the EEG datasets BCI-IV 2a, MAMEM II, and BCI Challenge. SS and SA denote subject-specific and subject-agnostic training of the native CBraMod model, respectively. \ding{100} indicates that the CBraMod backbone weights are frozen during training. We report accuracy for BCI-IV 2a and SSVEP and AUC for ERN, averaged over 5 runs. The last row (shaded) shows the standard LAtte model without any foundation model component for reference. The best result per dataset is highlighted in bold.
}
    \vspace{0.1em}
    \begin{small}
    \begin{tabular}{l@{\,+\,}l|ccc}
    \toprule
    \multicolumn{2}{l|}{\textbf{Models}} & \multicolumn{1}{c}{\textbf{BCIC\,IV\,2a}} & \multicolumn{1}{c}{\textbf{MAMEM II}} & \multicolumn{1}{c}{\textbf{BCI Challenge}} \\ \midrule
    \multicolumn{2}{l|}{CBraMod SS}            & 45.22$\pm$1.26 & 40.45$\pm$2.12 & 68.49$\pm$1.17 \\
    \multicolumn{2}{l|}{CBraMod SA}            & 39.14$\pm$1.55 & 49.91$\pm$0.00 & 72.90$\pm$0.46 \\
    CBraMod           & IncShallow             & 40.93$\pm$2.41 & 45.47$\pm$1.63 & 70.77$\pm$0.14 \\
    CBraMod\ding{100} & IncShallow             & 30.01$\pm$0.06 & 24.00$\pm$0.00 & 53.52$\pm$0.01 \\
    CBraMod           & IncDeep                & 41.71$\pm$2.23 & 48.09$\pm$0.42 & 70.03$\pm$0.10 \\
    CBraMod\ding{100} & IncDeep                & 32.83$\pm$0.00 & 26.87$\pm$1.57 & 57.39$\pm$9.32 \\
    CBraMod           & LAtte                  & 63.33$\pm$1.38 & \underline{66.24$\pm$0.71} & 63.39 $\pm$ 5.34 \\
    CBraMod\ding{100} & LAtte                  & \underline{64.38$\pm$1.60} & 65.73$\pm$0.01 & 72.38 $\pm$ 1.69  \\
    \multicolumn{2}{l|}{InceptionTime} & 62.85$\pm$3.21    & 62.71$\pm$2.95     & \underline{73.55$\pm$5.08}    \\
    \rowcolor{lightgray}
    \multicolumn{2}{l|}{LAtte}  & \textbf{78.85$\pm$2.55} & \textbf{71.53$\pm$1.00} & \textbf{83.8$\pm$1.12}\\
    \bottomrule
    \end{tabular}
    \label{tab:FM_ablation}
    \end{small}
\end{table}

\paragraph{Results} All CBraMod backbones are initialized from the published pretrained checkpoints of~\citet{DBLP:conf/iclr/WangZLZJLL025}. We retune hyperparameters for each new model configuration and repeat all experiments five times with different random seeds. To prevent the fine-tuning process from overwriting the representations learned during pretraining, we additionally evaluate variants in which the CBraMod backbone is frozen; these are denoted with~\ding{100}. The original CBraMod model is evaluated under two protocols: subject-specific (SS), in which a separate model is trained per subject, and subject-agnostic (SA), in which a single model is trained across all subjects without any additional subject information. For all datasets, we patch the sequence length with zeros to match the required dimensions of CBraMod.

\Cref{tab:FM_ablation} presents the results. CBraMod, with its fixed patch size of 200, is unable to effectively leverage its pretraining data on these benchmarks, particularly on BCI-IV 2a and MAMEM II, where sequence lengths are comparable to or shorter than the patch size itself. Notably, the subject-agnostic protocol outperforms the subject-specific setting for the native CBraMod model on two of three datasets, most likely because the larger effective training set slows the rate at which universal pretrained patterns are overwritten during fine-tuning.

The InceptionTime configurations perform worse than the native CBraMod model in all cases, indicating that the features extracted by the foundation model backbone do not contain patterns that the more expressive Inception architecture can exploit for these tasks. This becomes especially apparent with the frozen backbone, where performance degrades dramatically, in some cases to near or below chance level, suggesting that the Inception layers are unable to extract meaningful discriminative information from the pretrained representations when they cannot be adapted.

For CBraMod + LAtte, the impact of the foundation model backbone on performance is less severe because the two branches operate in parallel rather than sequentially. This architectural choice allows the Lorentz Inception branch to continue extracting features directly from the raw EEG sequences, partially compensating for the limitations of the CBraMod representations. Nevertheless, overall performance remains below that of the standard LAtte model without any foundation model component. As the InceptionTime experiments demonstrate, the representations produced by CBraMod's fixed patch sizes are a poor fit for the short trial durations in these benchmarks. Additionally, the baseline correction mechanism in LAtte is hindered by a geometric mismatch: the CBraMod backbone produces Euclidean representations, whereas the Lorentz Inception block operates on the hyperboloid, making the subtraction between the two branches less meaningful than in the standard configuration, where both branches share the same geometric space.

\subsubsection{Component Contribution}
\label{apx:ablations_comp}
In \Cref{tab:ablation}, we analyze the contribution of each component in the LAtte architecture by removing one module at a time. The largest performance drop occurs when replacing the Lorentz manifold with Euclidean space. Here, accuracy decreases by 19.75\%, confirming the importance of a more expressive geometric representation. Removing the subject-dependent LoRA parameters also leads to a substantial drop by 12.11\%, as the model can no longer effectively distinguish subject-specific distributions, which is critical for cross-subject generalization. The decoder projection and cut/fill augmentation contribute to improved regularization and stable convergence, which is an essential property for small EEG datasets.

\begin{table}[t]
\centering
\caption{Component Analysis of LAtte. We compare the performance of the full LAtte model to versions without the main modules on MAMEM II. We report the accuracy and std over 5 runs.}
\vspace{0.2em}
\begin{tabular}{llc}
\toprule
\textbf{Model Variant} & \textbf{Accuracy} & $\Delta$ \\
\midrule
LAtte Full   & 71.53$\pm$1.00 & -\\
LAtte w/o Lorentz     & 57.40$\pm$6.53 & -19.75\% \\
LAtte w/o LoRA       & 62.87$\pm$0.51 & -12.11\% \\
LAtte w/o Proj    & 69.34$\pm$1.37 & -3.06\% \\
LAtte w/o Cut/Fill    & 69.55$\pm$0.71 & -2.77\% \\
\bottomrule
\end{tabular}
\label{tab:ablation}
\vspace{-10pt}
\end{table}

We notice a similar loss in performance in \Cref{tab:ablation} when removing the baseline comparator in both BCIC-IV 2a and BCIC-IV 2b datasets. To keep fairness we matched the parameter count in the single block to having both blocks. Replacing hyperbolic maxpool with a tangent-based approach was not as damaging to the performance but operating directly on the hyperboloid remains the faster and more stable approach as well as it avoids excessive logmap and expmap operations.  
\begin{table}[t]
\centering
\caption{Additional Component Analysis of LAtte. We compare the performance of the full LAtte model to versions without the main modules on BCIC-IV 2a and BCIC-IV 2b.}
\vspace{0.2em}
\begin{tabular}{lcccc}
\toprule
\textbf{Model Variant}              & \textbf{BCIC-IV 2a} & $\Delta$   & \textbf{BCIC-IV 2b} & $\Delta$ \\
\midrule
LAtte Full                          & 78.85$\pm$2.55   & -         & 88.87$\pm$0.17   & - \\
LAtte w/o Baseline Block            & 72.90$\pm$1.67   & -7.54\%   & 84.50$\pm$0.47   & -4.9\% \\
LAtte w/o Hyperbolic MaxPool        & 77.01$\pm$1.34   & -2.33\%   & 87.39$\pm$0.48   & -1.66\% \\
LAtte w/o LB-LoRA                   & 77.21$\pm$2.13   & -2.1\%    & 86.29$\pm$0.30   & -2.9\% \\
\bottomrule
\end{tabular}
\label{tab:ablation_2}
\vspace{-10pt}
\end{table}

\subsubsection{Alternative LOSO Approaches}
\label{apx:loso_alt}
Our model architecture relies on subject IDs to model distributional shifts in the input data and bias the embeddings to subject-specific information. However, this would weaken the model's ability to generalize to unseen subjects since we do not have trained LoRA weights for the particular input distribution. In our methodology, we propose the use of a synthetic generalization subject ID that is stochastically injected into the training data with a certain probability. This is not the only available option as we are also able to use the base initialized LoRAs, which act as an identity on the input embedding, or use the LoRA's of the "most similar" other subject. We compare these three strategies below.

\paragraph{Baseline (fresh adapter)}
As the lower bound, the held-out subject's adapter slot is used as-is.
Because our LoRA weights are initialised with $\mathbf{R}{=}\mathbf{0}$ and $0$ boost velocity, the
adapter acts as an identity at the start of training and remains at its
randomly initialised state throughout. This means that the network sees the
unseen subject without any subject-specific conditioning.

\paragraph{Slot-0 generalisation}
During training, each sample's subject index is replaced with slot~0
(a reserved unknown subject slot) independently with probability~$p$.
This regularises slot~0 into a subject-agnostic generalised adapter.
At test time, the held-out subject's samples are all routed through this
trained slot~0.
Subject dropout is a training-time intervention: it does not require access
to the test subject's data, and the routing decision (slot~0) is fixed
before training begins.

\paragraph{Nearest subject routing}
An alternative is to route the held-out subject's test samples through the
adapter of the most similar trained subject.
We examine two proximity metrics.

\begin{itemize}
  \item \textbf{Lorentz-embedding distance}
    For each trained subject we compute the Lorentz centroid of the
    subject's embeddings produced by the live encoder on the training set,
    and similarly for the held-out subject on the test set.
    Proximity is measured as the geodesic distance on the hyperboloid
    $\mathcal{L}^n_k$ between these centroids.
    Because the centroids depend on the current encoder state, they are
    recomputed every $N$ epochs, making this a dynamic, model-aware metric.

  \item \textbf{SPD covariance distance}
    Subjects are compared via the affine-invariant Riemannian distance on
    average channel covariance matrices~\citep{pennec2006riemannian}:
    \begin{equation}
      d_\mathrm{SPD}(A,B)
        = \bigl\|\log\!\bigl(\mathbf{A}^{-1/2}\mathbf{B}\mathbf{A}^{-1/2}\bigr)\bigr\|_F.
      \label{eq:spd_dist}
    \end{equation}
    This metric is model-free and computed once per fold from the raw EEG
    signals, it is not sensitive to encoder evolution during training and
    serves as a strong geometry-agnostic baseline for subject similarity.
\end{itemize}

Unlike subject dropout, nearest-subject routing does not alter the training procedure but requires a pass over the held-out subject's test data to compute the proximity score.
The two routing variants thus differ in whether subject similarity is measured in the learned hyperbolic embedding space (model-dependent and dynamic) or in the signal covariance space (model-free and static).

\begin{table}[t]
\centering
\caption{Alternative LOSO approaches for the BCIC-2A dataset}
\vspace{0.2em}
\begin{tabular}{lc}
\toprule
\textbf{Model Variant} & \textbf{Accuracy} \\
\midrule
LAtte + Identity LoRA               & 44.22$\pm$4.56 \\
LAtte + Embedding Distance          & 57.32$\pm$1.51 \\
LAtte + SPD Covariance Metric       & 56.69$\pm$2.22 \\
LAtte + Slot-0 generalization       & 65.32$\pm$0.71 \\
\bottomrule
\end{tabular}
\label{tab:ablation_3}
\vspace{-10pt}
\end{table}

As we can see in \cref{tab:ablation_3}, the alternative approaches to the generalization slot lead to a very large degradation in performance. In the case of the nearest subject routing, the difference between both metrics is not very prominent, this is probably due to the model training being unaffected by either choice which limits the performance to simple LoRA selections which might not deviate too much.

%% file: figures/model_figure.tex
\definecolor{iceblue}{RGB}{173, 216, 230} 

\tikzset{
    processorblock/.style={rectangle, draw=black!70, fill=gray!3.5, thick, minimum height=6cm, minimum width=3.7cm, align=center}, 
    encoder/.style={rectangle, draw=black!70, fill=gray!3.5, thick, minimum height=6cm, minimum width=5.4cm, align=center}, 
        inc/.style={rectangle, rounded corners, draw=black!70,fill=blue!20, thick, minimum height=2cm, minimum width=1cm, align=center},
        norm/.style={rectangle, rounded corners, draw=black!70,fill=orange!25, thick, minimum height=3cm, minimum width=1cm, align=center},
        attention/.style={rectangle, rounded corners,draw=orange!70, fill=orange!20, thick, minimum height=5cm, minimum width=1cm, align=center},
    decoder/.style={rectangle, draw=black!70, fill=gray!3.5, thick, minimum height=6cm, minimum width=5.5cm, align=center}, 
        lcf/.style={rectangle, rounded corners, draw=black!70, fill=iceblue!20, thick, minimum height=4cm, minimum width=1cm, align=center},
        pd/.style={rectangle, rounded corners,draw=black!70, fill=blue!20, thick, minimum height=4cm, minimum width=1cm, align=center},
        processor/.style={rectangle, rounded corners,draw=black!70, fill=blue!20, thick, minimum height=4cm, minimum width=1cm, align=center},
        patching/.style={rectangle, rounded corners,draw=black!70, fill=gray!15, thick, minimum height=3cm, minimum width=1cm, align=center}, 
    arrow/.style={->, thick, >=stealth},
    line/.style={thick, draw=black},
}
\def\modgab{0.7cm}
\def\blockgap{0.5cm}

\begin{figure}
  \centering
  \resizebox{\linewidth}{!}{%
    \begin{tikzpicture}[node distance=2cm, auto]

    \node[processorblock, anchor=west] at (2,0) (procb) {}; 
    \node[xshift=-0.1cm, anchor=north east] at (procb.north east) {Processor};

    \node[left=2cm of procb, yshift=-2.5cm,
          label={Subject ID ($X^{\text{id}}$)}] (ID) {\includegraphics[width=0.15\linewidth]{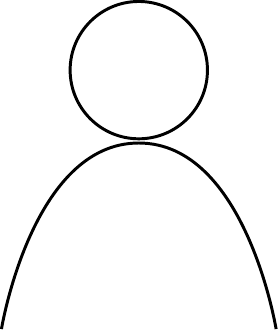}};
    
    \node[processor, right=\blockgap of procb.west] (proc) {\rotatebox{90}{Processor}};
    \node[patching, right=\modgab of proc] (patch) {\rotatebox{90}{Patching}};

    \coordinate (eeg_ref) at ($(procb.west)+(-1cm,2cm)$);
    \node[draw, inner sep=1mm, rounded corners, left=0.5cm of eeg_ref, yshift=-0.5cm, fill=white] (data1) {\includegraphics[width=0.2\linewidth, height=0.1\linewidth]{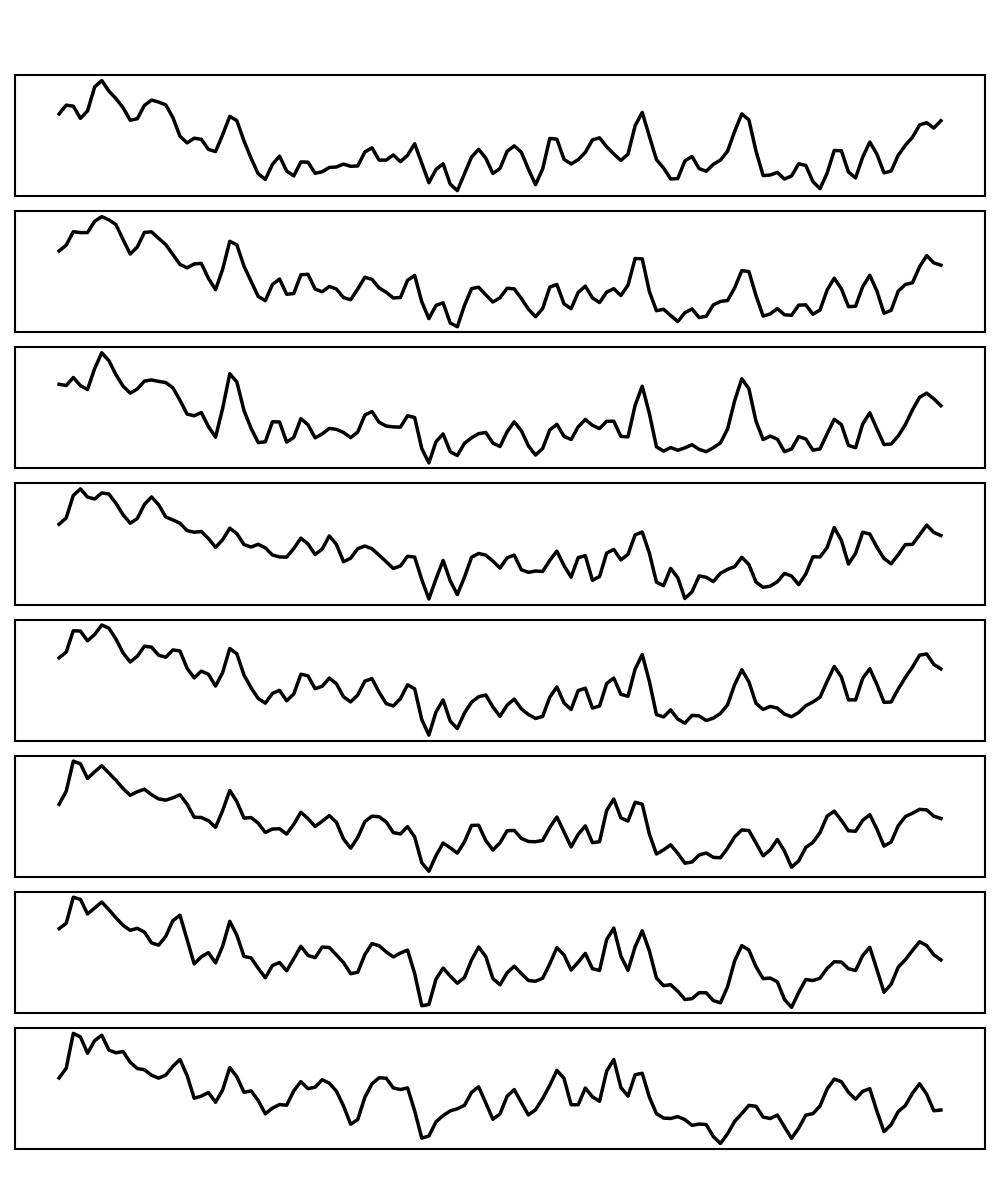}};
    \node[draw, inner sep=1mm, rounded corners, left=0.75cm of eeg_ref, yshift=-0.25cm, fill=white] (data2) {\includegraphics[width=0.2\linewidth, height=0.1\linewidth]{figures/ts.png}};
    \node[draw, inner sep=1mm, rounded corners, left=1cm of eeg_ref, fill=white] (data3) {\includegraphics[width=0.2\linewidth, height=0.1\linewidth]{figures/ts.png}};
    \node[draw, inner sep=1mm, rounded corners, left=1.25cm of eeg_ref, yshift=0.25cm, 
          label={EEG Sequences ($X^{\text{eeg}}$)}, fill=white] (data4) {\includegraphics[width=0.2\linewidth, height=0.1\linewidth]{figures/ts.png}};

    \draw[arrow] (data1.east) -- ++(0.5,0) coordinate (eeg_x) -- (eeg_x |- proc.west) -- (proc.west);
    
    \draw[arrow] (ID.east)  -- ++(1,0) coordinate (eeg_x) -- (eeg_x |- proc.west) -- (proc.west);
    
    \node[encoder, right=-0.2mm of procb] (enc) {};
    \node[xshift=-0.1cm, anchor=north east] at (enc.north east) {Encoder};
    
    \node[inc, anchor=north west] (inception) at ($(enc.north west) + (\blockgap, -5mm)$) {\rotatebox{90}{Inception}};
    \node[inc, anchor=south west] (baseline) at ($(enc.south west) + (\blockgap , 5mm)$) {\rotatebox{90}{Baseline}};
    
    \node[norm] (norm) at ($(inception.east)!0.5!(baseline.east) + (1.2cm,0)$) {\rotatebox{90}{Layernorm}};
    \node[attention, right=\modgab of norm] (att) {\rotatebox{90}{LorentzAtt}};
    
    \node[decoder, right=-0.2mm of enc] (dec) {};
    \node[anchor=north east] at (dec.north east) {Decoder};
    \node[patching, right=\blockgap of enc] (centroid) {\rotatebox{90}{CentroidUnpatch}};
    \node[lcf, right=\modgab of centroid] (lfc) {\rotatebox{90}{\ding{100} LorentzFC \ding{100}}};
    \node[pd, right=\modgab of lfc] (pd) {\rotatebox{90}{PrototypeDecoder}};

    \def\blockwidth{1cm}
    \def\blockheight{1cm}

    \coordinate (middlecoord) at ($(pd.east)+(1.5cm,0)$);

    \node[draw, fill=green!20, xshift=0.5cm, minimum width=\blockwidth, minimum height=\blockheight] (block4) 
        at ($(middlecoord)+(0,1.5*\blockheight)$) {};
    \node[draw, fill=red!20, xshift=0.5cm, minimum width=\blockwidth, minimum height=\blockheight] (block3) 
        at ($(middlecoord)+(0,0.5*\blockheight)$) {};
    \node[draw, fill=green!20, xshift=0.5cm, minimum width=\blockwidth, minimum height=\blockheight] (block2) 
        at ($(middlecoord)+(0,-0.5*\blockheight)$) {};
    \node[draw, fill=green!20, xshift=0.5cm, minimum width=\blockwidth, minimum height=\blockheight] (block1) 
        at ($(middlecoord)+(0,-1.5*\blockheight)$) {};

    \node[xshift=2.5cm, anchor=north east] at (dec.north east) {Classification};
    
    \coordinate (mid_after) at ($0.5*(inception) + 0.5*(baseline) + (0.7cm,0)$);
    \coordinate (mid_before) at ($0.5*(inception) + 0.5*(baseline) - (0.7cm,0)$);
    \coordinate (below_baseline) at ($(baseline) - (0cm,1.25)$);
    
    \draw[arrow] (proc.east) -- (patch.west);
    \draw[arrow] (norm.east) -- (att.west);
    \draw[line] (inception.east) -| (mid_after);
    \draw[line] (baseline.east) -| (mid_after);
    \draw[arrow] (mid_after) -- (norm.west);
    \draw[line] (patch.east) -- (mid_before.west);
    \draw[arrow] (mid_before.north) |- (inception.west);
    \draw[arrow] (mid_before.south) |- (baseline.west);
    
    \draw[arrow] (att) -- (centroid);
    \draw[arrow] (centroid) -- (lfc);
    \draw[arrow] (lfc.east) -- (pd.west);
    \draw[arrow] (pd.east) -- (middlecoord.west);

    \draw[arrow] (ID.south) -- ++(0,-0.7cm) coordinate (id_below) -- (id_below -| lfc.south) -- (lfc.south);

    \node[circle, draw, fill=white, inner sep=0.4pt] at (mid_after) {$-$};

    \end{tikzpicture}
    }
    \caption{LAtte architecture overview}
    \label{fig:latte}
\end{figure}